\RequirePackage{tikz}
\documentclass[sn-basic]{sn-jnl}

\usepackage{array}
\usepackage{longtable}
\usepackage{supertabular}
\usepackage{booktabs}
\usepackage{pgfplots}
\usepackage{pgfplotstable}
\usepackage{tikz}
\usepackage{multirow}
\usepackage{pgfplots}
\usepackage{tikz}
\usepackage{xspace}

\usepackage{pdflscape}
\usepackage{fancyhdr}

\usepackage{pgfplotstable}
\pgfplotsset{compat=1.14}
\definecolor{findOptimalPartition}{HTML}{696969}
\definecolor{storeClusterComponent}{HTML}{808080}
\definecolor{dbscan}{HTML}{BEBEBE}
\definecolor{constructCluster}{HTML}{DCDCDC}

\fancypagestyle{mylandscape}{
\fancyhf{} 
\fancyfoot{
\makebox[\textwidth][r]{
  \rlap{\hspace{.75cm}
    \smash{
      \raisebox{4.87in}{
        \rotatebox{90}{\thepage}}}}}}
}

\jyear{2021}%

\theoremstyle{thmstyleone}%
\theoremstyle{thmstyletwo}%

\theoremstyle{thmstylethree}%

\newcounter{notecounter}
\newcommand{\enotesoff}{\long\gdef\enote##1##2{}}

\enotesoff

\def\tailone{Glot500-only\xspace}
\def\tailtwo{tail\xspace}
\def\Tailtwo{Tail\xspace}

\begin{document}

\title[Article Title]{Taxi1500: A Multilingual Dataset for Text Classification in 1500 Languages}


\author*[1,2]{\fnm{Chunlan} \sur{Ma}}\email{machunlan@cis.lmu.de}

\author[1,2]{\fnm{Ayyoob} \sur{Imani}}\email{ayyoob@cis.lmu.de}

\author[1,2]{\fnm{Haotian} \sur{Ye}}\email{yehao@cis.lmu.de}
\author[1]{\fnm{Renhao} \sur{Pei}}\email{R.Pei@campus.lmu.de}
\author[3]{\fnm{Ehsaneddin} \sur{Asgari}}\email{asgari@berkeley.edu}
\author[1,2]{\fnm{Hinrich} \sur{Schütze}}\email{inquiries@cislmu.org}

\affil[1]{\orgdiv{CIS, LMU Munich}, \orgaddress{\country{Germany}}}
\affil[2]{\orgdiv{Munich Center for Machine Learning (MCML)}}
\affil[3]{\orgdiv{University of California, Berkeley}, \orgaddress{\country{USA}}}





\abstract{While natural language processing tools have been developed extensively for some of the world's languages, a significant portion of the world's over 7000 languages are still neglected. One reason for this is that evaluation datasets do not yet cover a wide range of languages, including low-resource and endangered ones. We aim to address this issue by creating a text classification dataset encompassing a large number of languages, many of which currently have little to no annotated data available.
We leverage parallel translations of the Bible to construct such a dataset by first developing applicable topics and employing a crowdsourcing tool to collect annotated data. By annotating the English side of the data and projecting the labels onto other languages through aligned verses, we generate text classification datasets for more than 1500 languages. We extensively benchmark several existing multilingual language models using our dataset. To facilitate the advancement of research in this area, we release 670 language in 1430 editions at the time of publishing. Our dataset and code are available at: \url{https://github.com/cisnlp/Taxi1500}.
}




\maketitle

\enote{hs}{you should explain what an abbreviation stands
  for (although there are excdeptoins). example: CLWE}

  \enote{hs}{in English punctuation, you usually have to leave
  space between a word and a foloowing opening
  parenthesis. please insert space weher eyou haven't
  already done so}
  \enote{cl}{punctuation fixed}

\enote{hs}{please remove space between footnote number
  (e.g., $^1$) and the preceding word}
\enote{cl}{footnote number space removed}

\enote{hs}{instead of
``Thus, \citep{schwenk-li-2018-corpus} sample the same
  number'', you need to write:
``Thus, \citet{schwenk-li-2018-corpus} sample the same
  number''.
  please correct that for all cites}
\enote{cl}{citation fixed}

\enote{hs}{
  instead of
  ``based on
their families \cite{hammarstrom2015glottolog}'', you need
to write:
  ``based on
their families \citep{hammarstrom2015glottolog}''}
\enote{cl}{punctuation fixed}

\enote{hs}{``in language'' vs ``in-language'': please use
  ``in-language'' consistently}
\enote{cl}{in-language consistency fixed}

\section{Introduction}\label{sec1}

Language inequality is a real issue in the world today as 
minority languages are under-represented and often excluded from language technologies \citep{joshi-etal-2020-state}.
The lack of technological support 
for minority languages in communities around the globe has a significant impact on the experience of their users and is commonly a cause for virtual barriers such as 
the \textit{digital divide}.\footnote{labs.theguardian.com/digital-language-divide}
Recent development in language technologies has brought 
about a surge of multilingual pre-trained language models (MPLMs), such as the multilingual BERT (mBERT) \citep{DBLP:journals/corr/abs-1810-04805}, XLM-R \citep{conneau-etal-2020-unsupervised}, and the more recently proposed SERENGETI \citep{adebara2022serengeti} and Glot500-m \citep{imani-etal-2023-glot500}, both of which support around 500 languages.
While the number of supported languages in the newest MPLMs keeps increasing, we are still unable to quantify the performance on most low-resource languages.
We believe that a major cause for why many low-resource languages are still neglected lies in the lack of evaluation datasets for such languages.
For example, MPLMs like mBERT and XLM-R are evaluated for many fewer languages than they are pretrained for because of the limited availability of languages in most benchmark datasets.

Most existing multilingual benchmarks, such as XNLI \citep{lewis-etal-2020-mlqa} and MLQA \citep{DBLP:journals/corr/abs-1909-04761}, rely on translating monolingual benchmarks, as opposed to collecting data from scratch.
This approach involves the translation of monolingual data either through machine translation or with the assistance of human professionals.
However, machine translation has limitations in terms of the number of languages that can be effectively handled, which depends on the supported languages of the machine translation system, while the quality of translations is also not guaranteed. On the other hand, human translation yields high-quality results but is accompanied by significant costs.


As a solution, we propose a dataset that covers more than 1500 languages. 
We use translations of the Bible as our source and develop
classification topics (i.e., classes) that are general
enough so as to apply to many verses and are at the same
time not overly abstract.
We obtain annotations for the English verses using crowdsourcing.
Because the Bible is aligned at the verse level, we can easily project annotations from the English side to all other languages.
We attempt to ensure the quality of our annotated
data, including by measuring inter-annotator agreement.
We name our dataset \textit{Taxi1500}.
As a case study, we evaluate three MPLMs (mBERT, XLM-R and Glot500-m) on Taxi1500. Our results suggest that Taxi1500 successfully demonstrates the multilingual generalizability of different MPLMs.

\section{Related Works}

\subsection{Multilingual datasets}

To date, most datasets that can be used for multilingual
task evaluation \citep{pan-etal-2017-cross,
  DBLP:journals/corr/abs-1909-04761, de2021universal} contain no more
than a few hundred languages, a small number compared to the
world's 7000 languages. In this section, we  give an
overview of existing state-of-the-art multilingual datasets.

\subsubsection*{The Universal Dependencies Treebanks}
Universal Dependencies (UD)
v2\footnote{\url{https://lindat.mff.cuni.cz/repository/xmlui/handle/11234/1-3687}}
is an evergrowing multilingual treebank collection, covering
90 languages and 17 tags. UD collects data from an evolution
of (universal) Stanford dependencies \citep{de2014universal},
Google universal part-of-speech tags
\citep{petrov-etal-2012-universal}, and the Interset
interlingua for morphosyntactic tagsets \citep{inbook}. UD
is often used as a  POS tagging component (representing structured prediction)
in multilingual benchmarks such
as XTREME \citep{DBLP:journals/corr/abs-2003-11080}. 

\subsubsection*{Wikiann}
\citet{pan-etal-2017-cross} develop a cross-lingual named entity tagging dataset in 282 languages based on articles from Wikipedia. The framework extracts name tags through cross-lingual and anchor links, self-training, and data selection methods and links them to an English Wikipedia Knowledge Base. Wikiann is recently used for the structured named entity prediction task for multilingual benchmarks such as XTREME \citep{DBLP:journals/corr/abs-2003-11080}.

\subsubsection*{Tatoeba}
Tatoeba \footnote{\url{https://tatoeba.org/eng/}} is a community-supported collection of English sentences and translations into more than 300 languages. The number of translations updates every Saturday. \citet{artetxe-schwenk-2019-massively} extract a dataset from Tatoeba with 1000 sentences in 112 languages. Multilingual benchmark XTREME \citep{DBLP:journals/corr/abs-2003-11080} collects this dataset as a task by calculating the cosine similarity to evaluate the performance of multilingual models.

\begin{table}
\centering
\begin{tabular}{c c c} 
 \hline
 Dataset & Languages & Tasks \\ [0.5ex] 
 \hline 
PAWS-X & 6 & Sentence-pair Classification \\
MLQA & 7 & Question Answering \\
MLDoc & 8 & Document Classification \\
XQuAD & 10 & Question Answering \\ 
XLIN & 15 & Sentence-pair Classification \\
XTREME* & 40 & MLQA, XQuAD, PAWS-X, XLIN, NER, POS\\
The Universal Dependency & 90 & POS\\
Wikiann & 258 & NER\\
Tatoeba & 300 & Machine Translation\\
 \hline 
\end{tabular}
\label{table: multilingual benchamrk}
\caption{Multilingual datasets and contained tasks. *XTREME contains the task from MLQA, XQuAD, PAWS-X, XLIN, and two additional tasks NER and POS.}
\end{table}

\subsubsection*{MLDoc}
Multilingual Document Classification \citep{lewis-etal-2020-mlqa} is a multilingual benchmark for document classification for eight languages: English, French, Spanish, Italian, German, Russian, Chinese, and Japanese. It uses data from the Reuters Corpus Volume 2 (RCV2) \citep{lewis2004rcv1}\footnote{\url{https://trec.nist.gov/data/reuters/reuters.html}}, a multilingual corpus with 487,000 news stories in thirteen languages (Dutch, French, German, Chinese, Japanese, Russian, Portuguese, Spanish, Latin American Spanish, Italian, Danish, Norwegian, and Swedish) that are manually classified into four groups: CCAT (Corporate/Industrial), ECAT (Economics), GCAT (Government/Social) and MCAT (Markets). MLDoc samples data with equal class priors to address the issue of data imbalance that previous research on RCV2 encountered. For example, \citet{klementiev2012inducing} define a subset of English and German portions from RCV2, which is used in several follow-up works, for instance, \citet{mogadala2016bilingual} extend the use of RCV2 to French and Spanish through transfer from English. These above-mentioned corpora obtain high accuracy during training but far lower accuracy during testing, which may be caused by the imbalanced distribution of each category. Thus, \citet{schwenk-li-2018-corpus} sample the same number of examples for each class and language from RCV2 instead of choosing a random subset to avoid an imbalanced dataset. The task of MLDoc is the same as RCV2, but with a balanced evaluation, the balanced dataset enables fair evaluation between different language models. 

\subsubsection*{XNLI}
The Cross-lingual Natural Language Inference Corpus(XNLI)
\citep{DBLP:journals/corr/abs-1909-04761} is a multilingual
evaluation benchmark that extends NLI to 15 languages,
namely English, French, Spanish, German, Greek, Bulgarian,
Russian, Turkish, Arabic, Vietnamese, Thai, Chinese, Hindi,
and two lower-resource languages Swahili and Urdu. XNLI
supports the evaluation of the NLI task:
two sentences are classified as
entailment, contradiction, or neither. XNLI comprises a
total of 112,500 annotated sentence pairs and follows the
same data collection procedure as the MultiNLI corpus
\citep{N18-1101}: 250 sentences are sampled from ten sources
of the Open American National Corpus: Face-To-Face,
Telephone, Government, 9/11, Letters, Oxford University
Press (OUP), Slate, Verbatim, and Government, and the
fiction novel Captain Blood. These sentences are then
translated into the other 14 languages using the \textit{One
  Hour Translation} crowdsourcing platform. This approach
enables language models to learn cross-lingual inference
ability by using pairs of premise and hypothese in
different languages.

\subsubsection*{MLQA}
MLQA \citep{lewis-etal-2020-mlqa} is a series of
multilingual extractive question-answering corpora available in seven
languages: English, Arabic, German, Vietnamese, Spanish,
Simplified Chinese, and Hindi. The resulting corpora have
over 12K instances in English and 5K in each other language,
with an average of four parallel sentences across languages per instance \citep{conneau-etal-2020-unsupervised}. The extraction process involves identifying paragraphs from Wikipedia articles that cover the same or similar topics in multiple languages. Subsequently, crowdsourcing is used to generate questions and answer spans from English paragraphs via the Amazon Mechanical Turk platform. The researchers employ professional translators to translate English questions into the target languages, and the translators then annotate answer spans within the corresponding paragraphs. MLQA leverages Wikipedia articles as the source due to Wikipedia's naturally parallel nature and large scale.


\subsubsection*{XQuAD}
XQuAD (Cross-lingual Question Answering Dataset) \citep{artetxe2019cross} is a multilingual benchmark dataset for cross-lingual question answering tasks. It involves the translation of 240 paragraphs and 1,190 question-answer pairs from the development subset of SQuAD v1.1 \citep{rajpurkar-etal-2016-squad} into ten languages, namely Spanish, German, Greek, Russian, Turkish, Arabic, Vietnamese, Thai, Chinese, and Hindi. Instead of collecting data from scratch, XQuAD translates data from SQuAD to avoid unanswerable questions. The researchers constructed XQuAD to mitigate the issue of superficial keyword matching problems that can arise in cross-lingual question answering tasks. The dataset is evaluated using both CLWE \citep{rajpurkar-etal-2016-squad} and a monolingual model.

\subsubsection*{XTREME}
The Cross-lingual TRansfer Evaluation of Multilingual
Encoders (XTREME)
benchmark\citep{DBLP:journals/corr/abs-2003-11080} is a
massively compiled multilingual benchmark compromising 40
languages and 9 tasks. As a state-of-the-art multilingual
benchmark, XTREME is designed based on available
multilingual corpora and their variable tasks. The nine
tasks are categorized into different categories, namely
classification, structured prediction, Question-Answering,
and Information Retrieval (shown in table \ref{table:
  multilingual benchamrk}). To obtain labeled data, the
authors utilized corpora such as XNLI, PAWS-X, Universal
Dependencies v2.5 (English for training and target language
test set for evaluation) and Wikiann (data selection). 

\subsection{Multilingual language models}

Multilingual models are large language models that cover
several different languages.  For instance, mBERT and XLM-R
are pre-trained on over 100 languages.
In this section, we discuss the four models that are
relevant to our Taxi1500 evaluation case study.


\subsubsection*{mBERT}
mBERT (Multilingual Bidirectional Encoder Representations
from Transformers) \citep{DBLP:journals/corr/abs-1810-04805}
is the multilingual variant of BERT, which follows the BERT
recipe closely. Similar to BERT, mBERT uses
the encoder architecture from Transformer, with Masked
Language Modeling (MLM) and Next Sentence Prediction (NSP)
as its training objectives. mBERT differs from BERT
primarily in its training data. While BERT is trained on
English Wikipedia and the Toronto Books Corpus, mBERT has a
training set from Wikipedia in 104 languages, with an uncased
version (102 languages) and a cased version (104
languages). Both versions have the same hyper-parameters: 12
layers, 768 hidden units per layer, 12 attention heads, a
110k shared WordPiece vocabulary, and 110M
parameters\footnote{\url{https://github.com/google-research/bert/blob}}. Normally,
the cased version is recommended because it fixes
normalization issues in many languages.

\subsubsection*{XLM}
XLM (Cross-Lingual Models)
\citep{DBLP:journals/corr/abs-1901-07291} is also a
BERT-based model and acquires multilingual ability by using
improved pre-training methods. XLM
uses three objectives, two of which are unsupervised and
merely require monolingual data. The two unsupervised tasks,
Causal Language Modeling (CLM) and Masked Language Modeling
(MLM) aim to learn cross-lingual representations. The
supervised objective
Translation Language Modeling (TLM)
relies on parallel corpora. 
TLM extends MLM from
BERT and replaces monolingual sentence pairs with parallel
multilingual sentence pairs. A sentence pair is concatenated
from a source language sentence and a target language
sentence. The words in the source and target sentence are
randomly masked. To predict the masked words, the model is
allowed to attend to the source language sentence or the
target language sentence, which enables the model to align
the source and target sentence. Therefore, the model obtains
cross-lingual ability with TLM modeling. As for the training
data, XLM crawls Wikipedia dumps as monolingual data for CLM
and MLM objectives. For TLM, the authors only use parallel
data containing English from different resources such as
MultiUN \citep{ziemski-etal-2016-united}, IIT Bombay corpus
\citep{kunchukuttan-etal-2018-iit} and EUbookshop
corpus. For the embeddings, fastBPE is applied to split
words into subword units. The authors release multiple
pre-trained versions of XLM, the most massively-multilingual
variant is XLM-R \citep{conneau-etal-2020-unsupervised}.

\subsubsection*{XLM-R}
XLM-R \citep{conneau-etal-2020-unsupervised} is an improved
version of XLM. Inspired by RoBERTa
\citep{DBLP:journals/corr/abs-1907-11692}. The authors claim
that mBERT and XLM are both undertrained. Therefore, they
pre-train XLM-R with larger model size and massive data
from Common Crawl in 104 languages, significantly boosting
the performance and outperforming mBERT. Compared with XLM,
XLM-R has a larger vocabulary size of 250K. Besides, the
training data scales from Wikipedia to a larger Common Crawl
corpus. The authors provide two XLM-R versions, XLM-R Base
(12 layers, 768 hidden units, 12 attention heads, 270M
parameters) and XLM-R Large (24 layers, 1024 hidden units,
16 attention heads, 550M parameters). XLM-R mitigates the
curse of multilinguality (i.e., it addresses the increased
need for parameters when the number of covered languages increases)
by increasing the model capacity.

\enote{hs}{LLM or PLM? make consistent}

\subsubsection*{Glot500-m}
Utilizing continuous pretraining based on XLM-R, Glot500-m\citep{imani-etal-2023-glot500}
was developed as a multilingual LLM on 500 languages. To
train the model,  the
authors collect and clean a corpus, Glot500c, that covers
more than 500 languages.
Glot500m is
evaluated on six  tasks, namely sentence retrieval
Tatoeba, sentence retrieval Bible, Taxi1500, text
classification, NER, POS and round-trip alignment. The
authors illustrate results with two language sets: head
languages (104 pretrained languages of XLM-R) and tail
languages (the remaining languages) and compare results with
XLM-R-Base and XLM-R-Large. They find that multilingual
LLMs' quality is not only influenced by a single isssue, but
is determined by several factors, including corpus, script and related languages.

\subsubsection*{SERENGETI}
SERENGETI \citep{adebara2022serengeti} is pretrained with
517 African languages and the 10 most spoken languages in the
world. It is an Electra \citep{clark2020electra} style model. To
obtain the training data, the authors collect a multi-domain
 multi-script corpus manually. The corpus includes religious
domain, news domain, government documents, health documents
and some data from existing corpora. SERENGETI is evaluated
on seven task clusters, containing NER, POS, phrase
chunking, news classification, sentiment classification and
topic classification and the results are provided as AfroNLU
benchmark. Their evaluation indicates that SERENGETI outperforms
XLM-R \citep{conneau-etal-2020-unsupervised}, KinyarBERT
\citep{Nzeyimana_2022}, AfriBERTA
\citep{ogueji-etal-2021-small} and Afro-XLMR
\citep{alabi-etal-2022-adapting} on 11 datasets with 82.27
average F1.

\subsection{The Parallel Bible Dataset}
\label{parallel Bible dataset}
In current NLP research, parallel corpora play a crucial
role as they serve as cross-lingual bridges, enabling the
processing and understanding of less known languages through
other languages.  In this study, we employ translations of
the Bible as the source of parallel data, utilizing both the Parallel Bible Corpus \citep{mayer-cysouw-2014-creating},
covering 1304 languages, as well as additional translations
collected from the web, resulting in total coverage of
1500+
languages. While there are other resources for parallel data
available, such as Europarl \citep{koehn-2005-europarl},
JW300 \citep{agic-vulic-2019-jw300}, and OPUS
\citep{TIEDEMANN12.463}, we have chosen to use translations
of the Bible due to the relatively larger number of
supported languages.


\subsubsection{PBC}
PBC consists of three main parts, namely the .txt files of
actual Bible texts, the .wordforms files that alphabetically
list all word forms in the texts, and the .mtx files
(word-by-verse matrices). Our work only uses the .txt
files. Every text file is one language version of the
Bible. Below we include several examples to illustrate the inner structure of the text file: every line contains an ID and the respective verse which is tokenized. The verse IDs are identical in different languages. When we use the parallel dataset, we can find the same verse in other languages with the help of the verse ID.

\enote{hs}{put the tabular environment into a proper table}
\enote{cl}{fixed}

\begin{table}[!]
\centering
\begin{tabular}{ll}
01001001&	In the beginning God created the heavens and the earth .\\
01001001&	Im Anfang erschuf Gott die Himmel und die Erde .\\
01001001&	Au commencement Dieu créa les cieux et la terre .
\end{tabular}
\label{tab: pbc_format}
\end{table}

\subsubsection{1000Langs}
The 1000Langs corpus contains the crawled data of 1500+
unique languages, which are sourced from multiple Bible
websites. The two main websites are \url{https://png.bible} and \url{https://ebible.org}.

\section{Dataset Creation}



\subsection{Principles of task design}
In designing and creating Taxi1500, we
exploit the nature of PBC (i.e., the fact that it
is a parallel corpus), but are at the same time
limited by some of its specifics.
In particular, we
were guided by two
considerations: cost efficiency and influence of the domain
of the Bible (i.e., religious text).

\bmhead{Cost efficiency} In order to lower the cost, we
exclude tasks that require the hiring of target language
experts for annotation.  Given that the PBC is
sentence-aligned, these include tasks such as NER and POS
tagging, which are based on word units.  We also exclude
tasks such as question answering, which require data
generation on the side of target language
experts. Therefore, the sentence classification task is
chosen as our task  to utilize the PBC
without knowledge of other languages. As the PBC is a
parallel text that is sentence aligned, we can easily
obtain the label of verses in other languages, as long as we
obtain annotations of English verses.

\bmhead{Bible domain} There are many kinds of classification
tasks.
We exclude sentiment and emotion classification because many
verses are objective descriptions of a state or event. In
the end, we select topic classification as the a
classification task for Taxi1500 because it can be naturally
applied to the Bible.

\subsection{Data annotation}
We describe our data annotation procedure in detail. Since
many low-resource languages only have a translated New
Testament, we use verses from the New Testament to build our dataset. In the first round of annotation, three annotators develop a few possible topics for a subset of the 
New Testament verses and decide on six final topics after
discussion: recommendation, faith, description, sin, grace,
and violence (see table \ref{tab:definition_verses}). We
select verses for which at least two of the three annotators agree.
We then remove verses that receive multiple topic
assignments and those that receive none.
The motivation is that removing ambiguous verses makes the
annotation task easier for annotators (which also reduces the
annotation cost). We submit the
remaining 1077 verses to Amazon
MTurk\footnote{www.mturk.com}, a crowdsourcing platform, for
annotation and specify the US as the annotators'
location. Each verse is eventually annotated ten times in
total. The final labels are selected by majority voting. In
the case of a tie, the final label is randomly selected from the majority labels.



Issues of annotation quality may arise if 1)
the task is confusing, and 2) the worker does not annotate
carefully.  To lessen the effects of the first problem, we
provide detailed guidelines and examples to the annotators.
Annotators are required to pass a qualification
test which is given prior to each annotation batch,
making sure they understand the task fully.
Additionally, we implement quality control in the form of a
performance threshold. Specifically, we construct
``pseudo gold standard'' data, i.e., labels derived from the
majority vote among all annotators, and compare them with
the annotations of each worker. The annotations of a
worker are rejected and the verses are republished for
a new round of annotation if the worker obtains an F1 of $<.4$.


We use Krippendorff's $\alpha$ (K-$\alpha$) to compute
inter-annotator agreement.  K-$\alpha$ is chosen because it
can handle missing annotations in the dataset since each
worker only annotates a subset of the verses.  Table
\ref{tab:k_alpha} shows K-$\alpha$ values for different
thresholds, i.e., the minimum votes for the majority label
required for a verse to be accepted.  We obtain K-$\alpha$ =
.44 on the entire dataset, which can be improved by raising
the threshold of required votes. But as Figure
\ref{fig:k_alpha} demonstrates, there is a clear tradeoff
between the number of accepted verses and K-$\alpha$, and
improving K-$\alpha$ would reduce the size of the dataset.
Furthermore, a slightly suboptimal K-$\alpha$ value is
not surprising considering that the topics of our task are
subjective, and as
\citep{price-etal-2020-six} points out, a low K-$\alpha$
does not necessarily signify low data quality.  We thus do
not remove any data by raising the required number of votes
and rely on the control measures described above to ensure
data quality.

\begin{table}
  \centering
  \small
  \begin{tabular}{p{2.8cm}|p{7.8cm}}
  \midrule
  \multicolumn{1}{c}{class} & \multicolumn{1}{c}{definition} \\
  \hline
  Recommendation & \texttt{An imperative statement which suggests to act or believe in certain ways.} \\
    \hline
  Faith & \texttt{Display of belief and love toward God, instructions on how to maintain faith, stories of faith and its consequences, etc.}\\
  \hline
  Description & \texttt{Describes a person, relationship, phenomenon, situation, etc. } \\
    \hline
  Sin & \texttt{Describes what is considered sin, stories of sinful people and sinful actions.}\\
    \hline
  Grace & \texttt{God’s love, blessing, and kindness towards humans.} \\
    \hline
  Violence & \texttt{Describes wars, conflict, threats, and torture; but also destructions of people, cities, and nations.} \\
  \hline
  \end{tabular}
  \vspace{0.2cm}
  \caption{
    \label{tab:definition_verses}
    Definitions of the six Taxi1500 classes
  }
\end{table}

\begin{table}
  \small
  \centering
  \setlength{\tabcolsep}{3pt}
  \begin{tabular}{c|ccccccc}
  \hline
  vote $\geq$ & 3    & 4    & 5    & 6    & 7    & 8    & 9     \\
  \hline
  num. verses & 1077 & 1055 & 941  & 755  & 563  & 388  & 233   \\
  \hline
  K-$\alpha$  & 0.44 & 0.44 & 0.48 & 0.55 & 0.63 & 0.73 & 0.83  \\
  \hline
  \end{tabular}
  \caption{
    \label{tab:k_alpha}
    The K-$\alpha$ value increases as we specify a higher threshold for the minimum number of votes of the majority topic. 3 is the lowest value here since we do not have any verses where the majority label has $<$ 3 votes.
  }
\end{table}

\pgfplotsset{every tick label/.append style={font=\tiny}}
\pgfplotsset{every axis/.append style={font=\small}}
\begin{figure}[h]
  \centering
  \begin{tikzpicture}
  \begin{axis}[
      width=0.8\columnwidth,
      height=4.5cm,
      xlabel={K-$\alpha$},
      ylabel={num. verses},
      xmin=0.4, xmax=0.85,
      ymin=180, ymax=1100,
      xtick={0.5,0.6,0.7,0.8},
      ytick={200,400,600,800,1000},
      xtick distance=1cm,
      legend pos=south west,
      ymajorgrids=true,
      grid style=dashed,
  ]
  \addplot[color=blue, mark=*] coordinates {(0.44,1077)(0.44,1055)(0.48,941)(0.55,755)(0.63,563)(0.73,388)(0.83,233)};
  \end{axis}
  \end{tikzpicture}
  \caption{\label{fig:k_alpha}
  Tradeoff between K-$\alpha$ and the number of verses. Each dot in the plot stands for a threshold of the required minimum votes $\in \{3,4,5,6,7,8,9\}$ for a verse to be accepted.
  }
\end{figure}
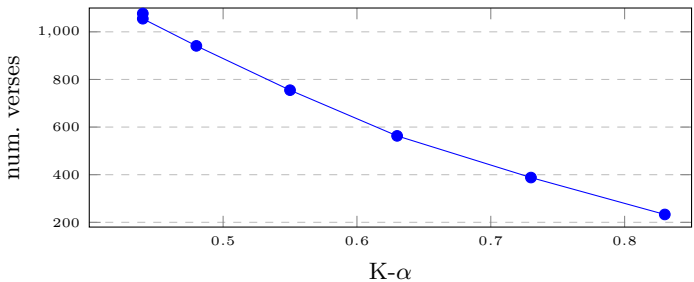

\section{The Dataset}
The final Taxi1500 dataset consists of 1077 verses categorized into six topics: 
faith, grace, sin, violence, description, and recommendation.
Table \ref{tab:example_verses} shows an overview of the topics with one 
example for each, as well as the number of verses of each topic in the 
English dataset.
\emph{Violence}, with 59 instances, is the smallest class and
\emph{recommendation}, with 281, is the largest.
Since some languages have incomplete translations of the New Testament 
and do not contain all of the 1077 verses, we exclude languages where 
the total number of annotated verses is less than 900. This leaves us 
with 1504 languages from 113 language families which are spread across 
the globe\footnote{family and geographical data from \url{glottolog.org}}. The dataset obtained for each of the 1504 languages is split into train, development, and test sets with a ratio of
80/10/10, with 860, 106, and 111 verses respectively\footnote{development and test sets have different sizes, because we split off train and development verses using their respective ratios and treat the rest as test verses}.

\begin{table}[ht]
  \centering
  \small
  \begin{tabular}{p{2.0cm}p{7.0cm}m{0.5cm}}
  \midrule
  \multicolumn{1}{c}{class} & \multicolumn{1}{c}{example} & \multicolumn{1}{c}{num. verses} \\
  \midrule
  recommendation & \texttt{If you love me , you will observe my commandments} & 281 \\
  [1ex]
  faith & \texttt{Most truly I say to you , whoever believes has everlasting life} & 260 \\
  [1ex]
  description & \texttt{There was a man of the Pharisees named Nicodemus , a ruler of the Jews} & 184 \\
  [1ex]
  sin & \texttt{That is why I said to you : You will die in your sins . For if you do not believe that I am the one , you will die in your sins .} & 153 \\
  [1ex]
  grace & \texttt{The Father loves the Son and has given all things into his hand} & 140 \\
  [1ex]
  violence & \texttt{He put James the brother of John to death by the sword} & 59 \\
  \hline
  \end{tabular}
  \vspace{0.3cm}
  \caption{
    \label{tab:example_verses}
    The table gives an example verse and the total number of verses in the crowdsourced English dataset for each class.
  }
\end{table}

\enote{hs}{is dishonoring really a sin? can you please find
  a clearer example of sin?}
\enote{cl}{example changed}

\begin{figure}
\centering
\includegraphics[width=0.49\textwidth]
{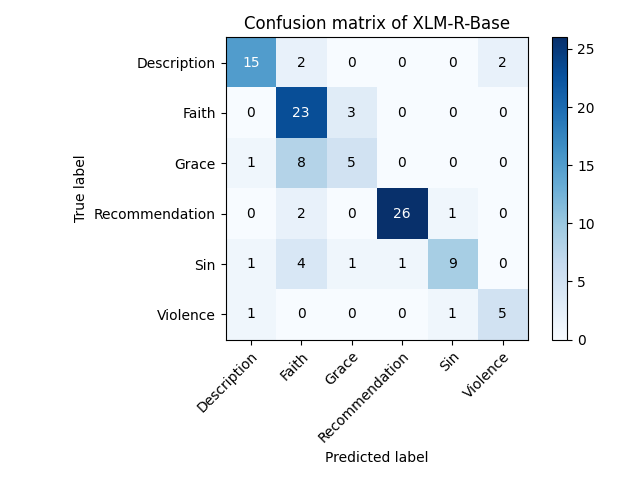}
\includegraphics[width=0.49\textwidth]
{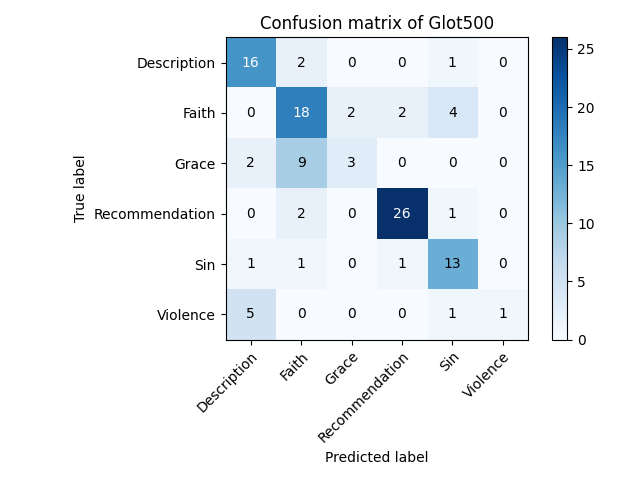}
\caption{Confusion matrices of five-fold cross validation of XLM-R-Base and Glot500.}
\label{fig:confusion matrix}
\end{figure}

\begin{table}
\centering
\resizebox{\textwidth}{!}{
\begin{tabular}{l|llllllllllllllllll}
\hline
verse.num & 1077 & 1076 & 1075 & 1074 & 1073 & 1072 & 1071 & 1070 & 1069 & 1067 & 1066 & 1065 \\
lan.num   & 1409 & 20   & 14   & 5    & 4    & 2    & 3    & 5    & 1    & 2    & 2    & 3    \\
\hline
verse.num & 1064 & 1063 & 1061 & 1060 & 1057 & 1056 & 1055 & 1054 & 1053 & 1051 & 1049 & 1048\\
lan.num & 3    & 1    & 2    & 3    & 1    & 2   & 3    & 1    & 1    & 1    & 1    & 3   \\
\hline
verse.num  & 1044 & 1042 & 1041 & 1039 & 1038 & 1034 & 1017 & 1006 & 1000 & 989  & 961  & 949   \\
lan.num     & 1    & 1    & 1    & 1    & 1    & 1    & 1    & 2    & 1    & 1    & 1    & 1   \\ 
\hline
\end{tabular}
}
\caption{
    \label{tab:language_verses}
    An overview of the number of verses of different languages, for example: 1049 of the languages have 1077 verses in the dataset.
  }
\end{table}


\enote{hs}{please reformat table tab:language\_verses so that
  most languages at the ``tail'' are consolidated into a few
  bins. E.g., you could have a bin 949--989 (with 3 members)}
\enote{cl}{I reformated the table and consolidated tail numbers, but it looks not nice because 949--989 or 1041-1049 are too wide comparing to the bins in the two lines above, I present the example there.}

To show more details of Taxi1500's topics, we present confusion matrices of five-fold cross-validation of XLM-R-Base and Glot500 in Figure \ref{fig:confusion matrix}. The matrices show that the topics Sin and Grace tend to be classified more frequently as other topics. This indicates that verses in Sin and Grace are more ambiguous to the models.

\enote{hs}{figure 3: please provide subheadings for the
  three subfgiures: head languages, glot500-only languages,
  tail languages}

\subsection{The corpus: Taxi1500-c}
To provide public access to our dataset, we have carefully selected uncopyrighted Bibles from the PBC and 1000Langs. We then compiled a corpus named Taxi1500-c, which includes all the Bibles that we can freely distribute. The current version available is Taxi1500-c v1.0.

\section{A Case Study for the Use of Taxi1500}
To illustrate its utility, we use Taxi1500 to evaluate four pre-trained multilingual models: mBERT, XLM-R-Base, XLM-R-Large, and Glot500-m. For a fair comparison, we split languages in our dataset into three sets, namely \textbf{head languages}, \textbf{\tailone languages}, and \textbf{\tailtwo languages}.
Head languages are languages that are in the pre-training data of all four models.
\tailone languages are languages that are only in the pre-training data of Glot500 and not the other three.
\Tailtwo languages include languages that are not in the pre-training data of any model.
Of the 1504 languages, there are 73 head languages, 250 \tailone languages, and 1149 \tailtwo languages.
We describe the detailed experiment setup in \ref{subsec:experiment_setup} and present the metrics on the test set in \ref{subsec:experiment_results}.

\subsection{Experimental Setup} \label{subsec:experiment_setup}

We conduct experiments on zero-shot transfer and on
in-language learning.
For all experiments, we select the best
checkpoint based on the validation loss and then report macro F1 score on the test set. We use the AdamW optimizer with learning rate $2e-5$ and batch size $\in \{2, 8, 16, 32\}$ and select the best result based on development set. All experiments are performed on a single GeForce GTX 1080Ti GPU.

\enote{hs}{do you select the best batch size based on dev? i
  don't think you say that?}
\enote{cl}{added}


\enote{hs}{what is a PMLM?}
\enote{cl}{changed to MPLM}

\bmhead{Zero-shot transfer} 
In zero-shot transfer,
we train (i.e., finetune) on the English training set and test on the test
set of the target language.

\bmhead{In-language learning}
In in-language learning, we train (i.e., finetune) on 
target language training data and test on the test set of the
target language.
We vary the size of the target language training set and
experiment with the 
following training set sizes:
$\{50, 100, 200, 400, 600, 860\}$. The training set size  860 corresponds to 
the full training set.
This allows us to investigate 
the effect of different 
amounts of training data.

\bmhead{Evaluation measure}
All results presented in this paper are macro-f1, which is chosed considering the imbalance of Taxi1500 dataset.

\enote{hs}{you need to describe the evaluation measure here:
  F1 (sorry if you already did that and I missed it)}
\enote{cl}{added}

\subsection{Results} \label{subsec:experiment_results}
In this section, we present experimental results on
zero-shot transfer, in-language learning,
analysis of the effect of training set size and analysis
based on language families.

\subsubsection{Zero-shot transfer}

\textbf{Baseline}
We conducted a Bag-of-Words (BOW) classification experiment
with our dataset and present the results as a baseline in
Appendix \ref{sec:appendix_zero_shot_results}. The
experiment revealed extremely low accuracy for BOW,
indicating that to classify verses in our dataset correctly,
the models must have access to a good semantic
representation.
The BOW representation does not
seem to be such a representation.

\begin{figure}[h!]
\centering
{\bfseries F1 of head languages}
\includegraphics[width=10cm, height=4cm]{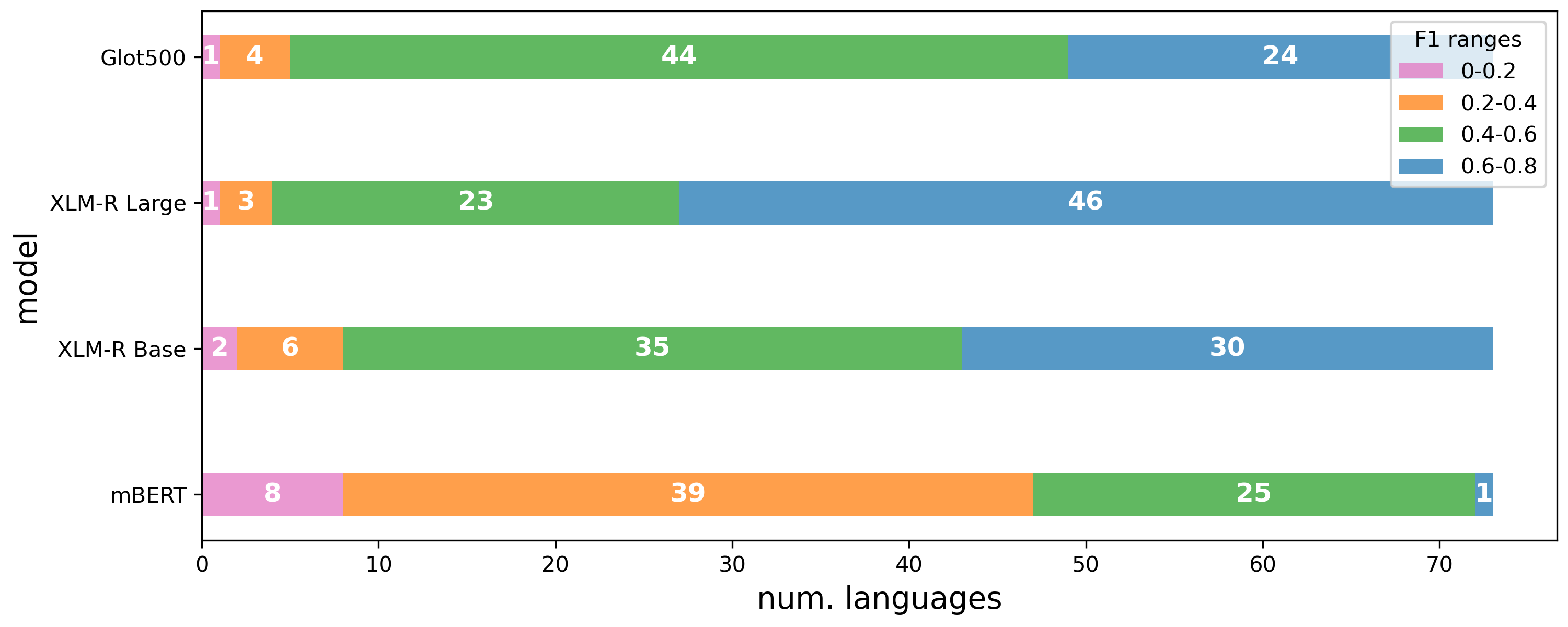}

{\bfseries F1 of head \tailone languages}
\includegraphics[width=10cm, height=4cm]{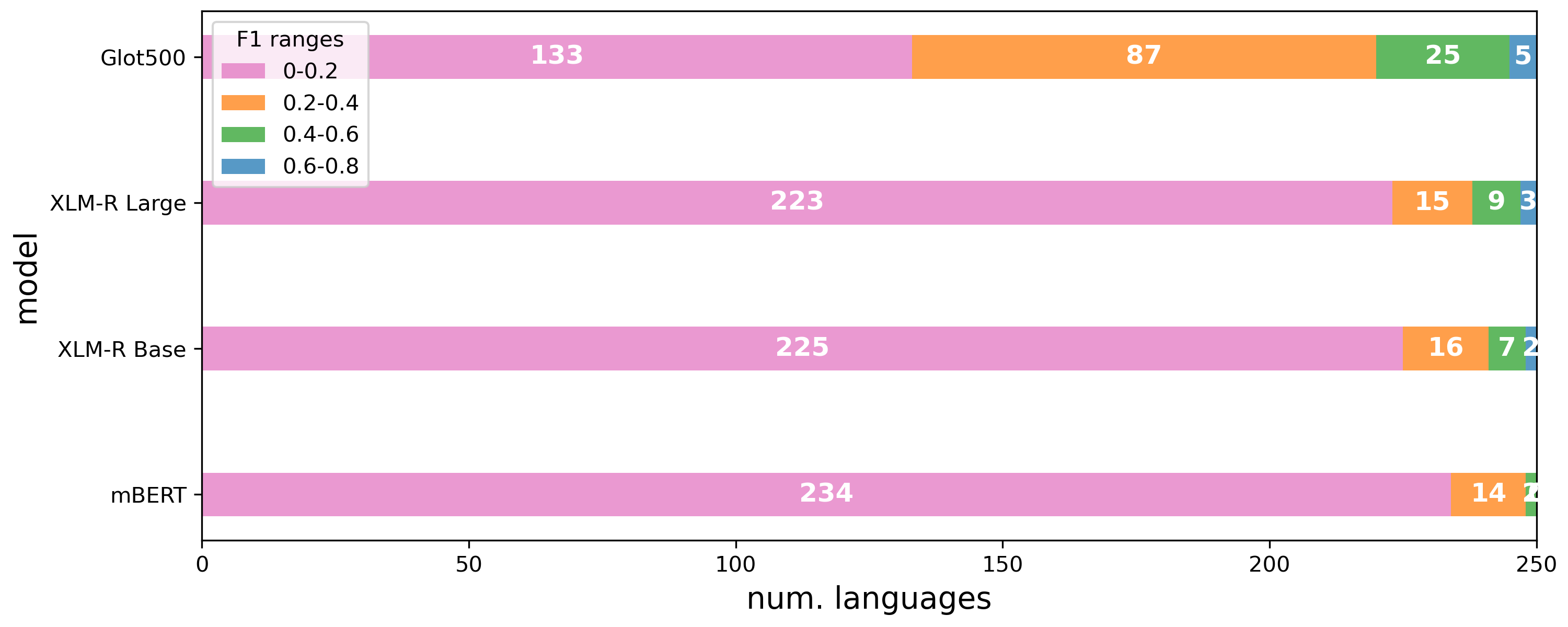}

{\bfseries F1 of head \tailtwo languages}
\includegraphics[width=10cm, height=4cm]{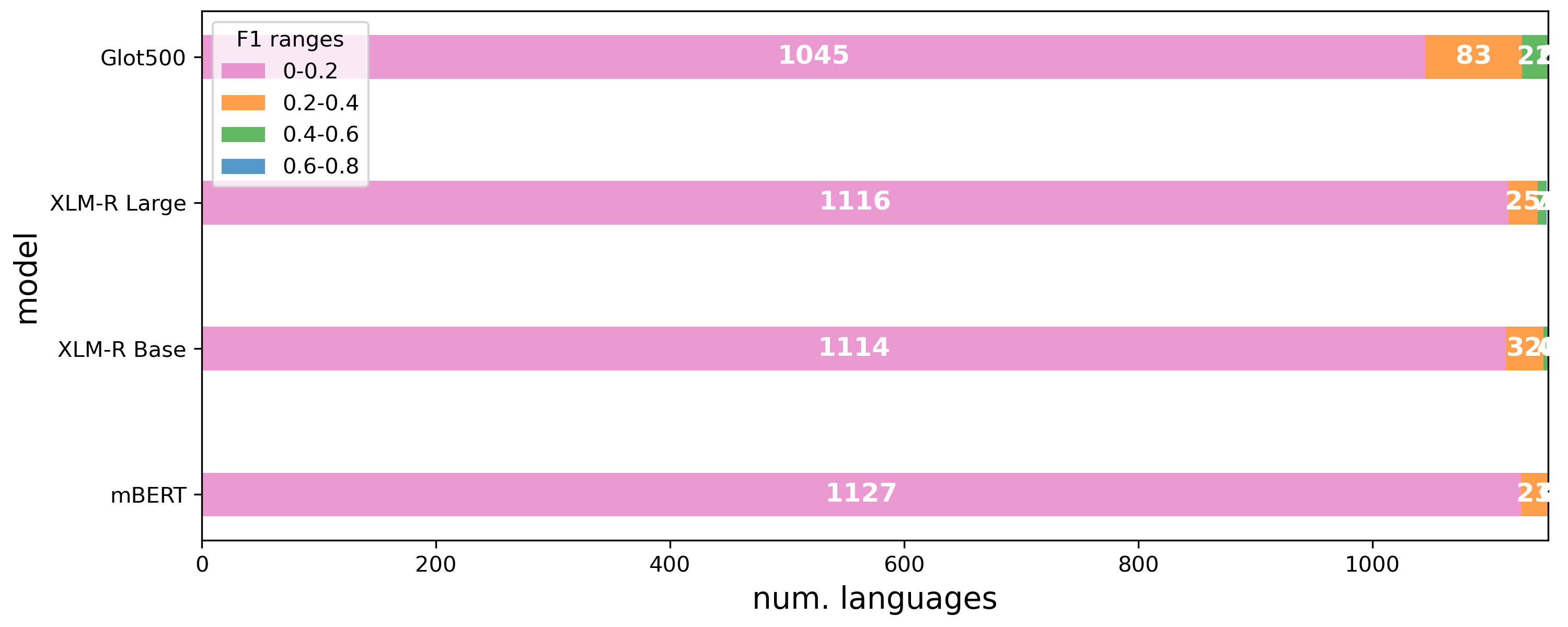}
\caption{Zero shot transfer learning: head languages (top), \tailone languages (middle) and \tailtwo languages (bottom). X-axis is the number of languages, y-axis presents four models. We split F1 scores into four ranges: 0-0.2, 0.2-0.4, 0.4-0.6 and 0.6-0.8.}
\label{fig:zeroshot_four_models}
\end{figure}

In figure \ref{fig:zeroshot_four_models}, we show the
results for 1504 languages, divided into three sets: head
languages (top), \tailone languages (middle), and \tailtwo
languages (bottom). On head languages, Glot500, XLM-L-B, and
XLM-R-L have 68, 65, and 69 languages within the high F1 range
(0.4-0.8), respectively, while mBERT only has 26 languages
within this range, indicating its worse performance. This
might be explained by a smaller pretraining data size of
mBert compared with the other three models. On \tailone
languages, Glot500 outperforms the other three models with
117 languages in the range of 0.2-0.8, whereas the other
three models have less than 30 languages within this
range. Because \tailone languages are in the pre-training
data of Glot500, we expect Glot500 to achieve better results
on these languages. On \tailtwo languages, Glot500
outperforms the other three models slightly with around 100
fewer languages in the range of 0-0.2. The reason might be that
a larger number of pre-training languages contributes to
higher performance for other \tailtwo languages from the same
family. The zero-shot transfer results
indicate that Taxi1500 can effectively demonstrate better
performance for models pretrained using more languages.

\enote{hs}{figures 5/6: the names are upside down!}

\subsubsection{In-language learning}

\begin{figure}[h!]
\centering
{\bfseries F1-differences of head languages}
\includegraphics[width=10cm, height=4cm]{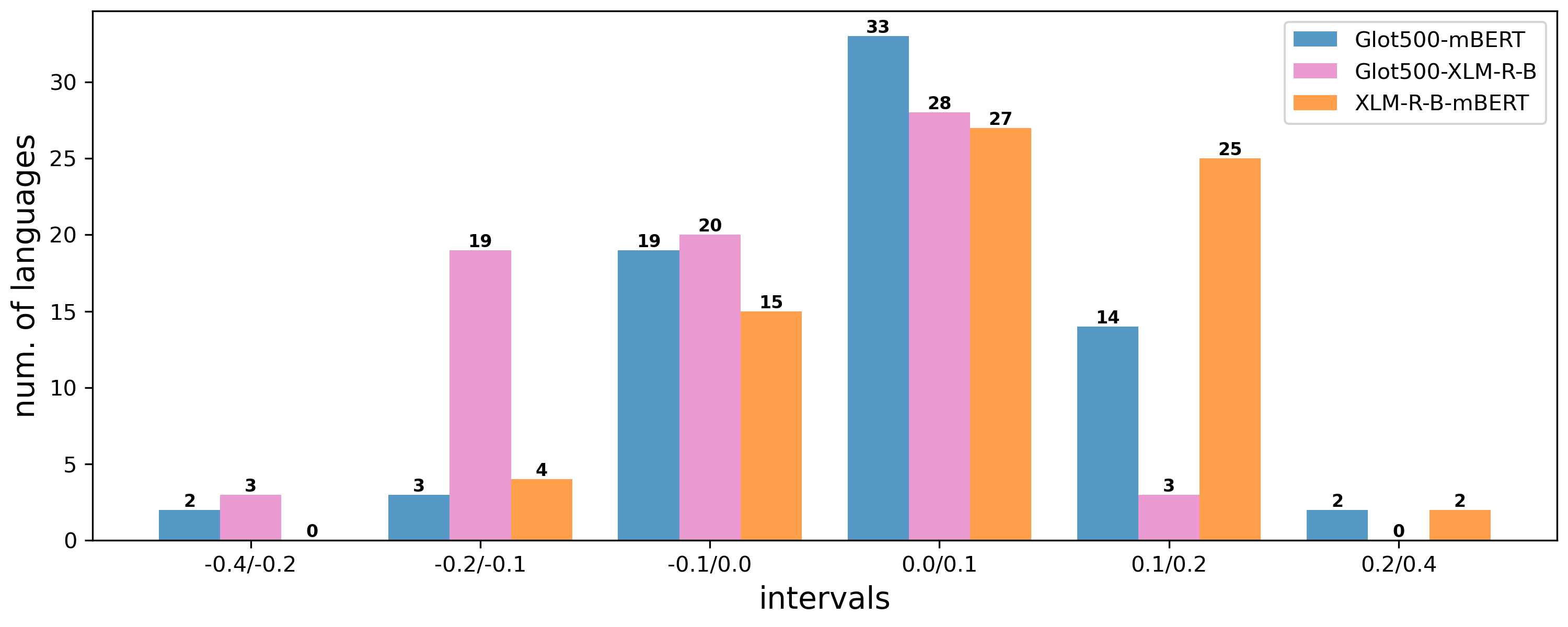}

{\bfseries F1-differences of \tailone languages}
\includegraphics[width=10cm, height=4cm]{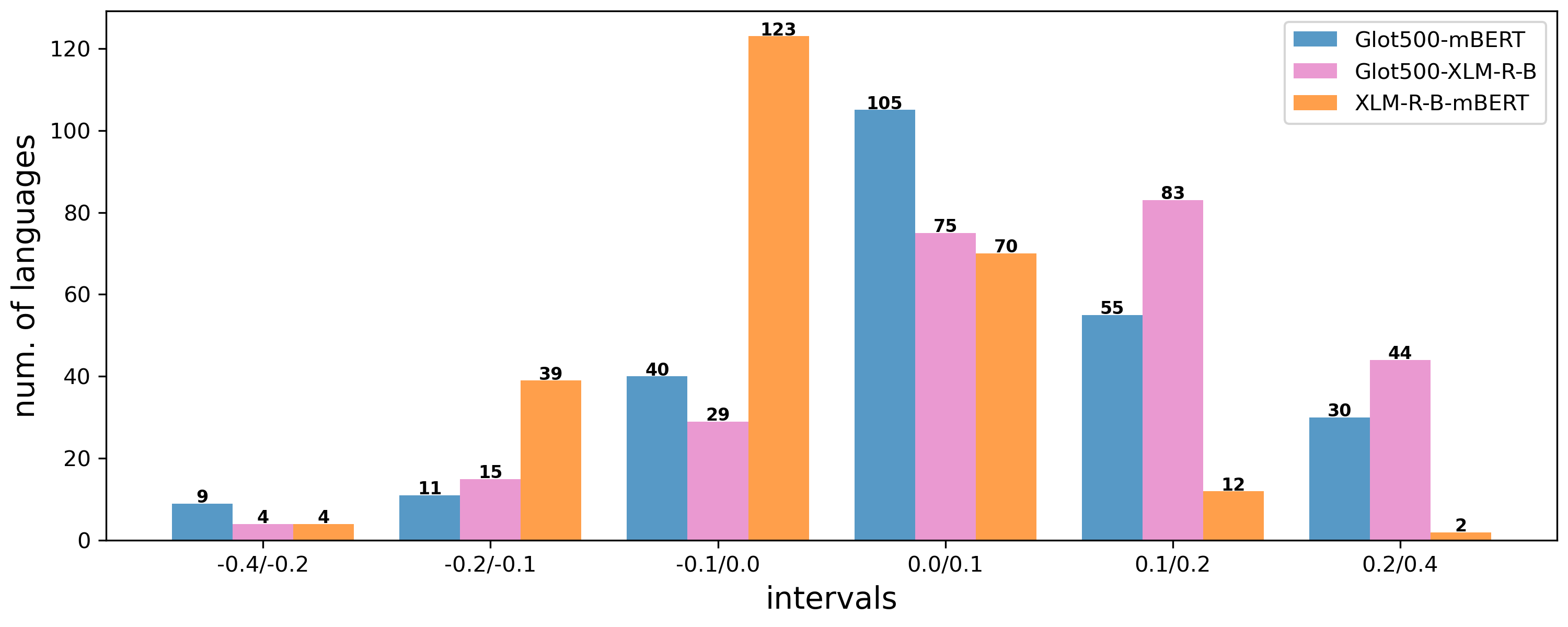}

{\bfseries F1-differences of \tailtwo languages}
\includegraphics[width=10cm, height=4cm]{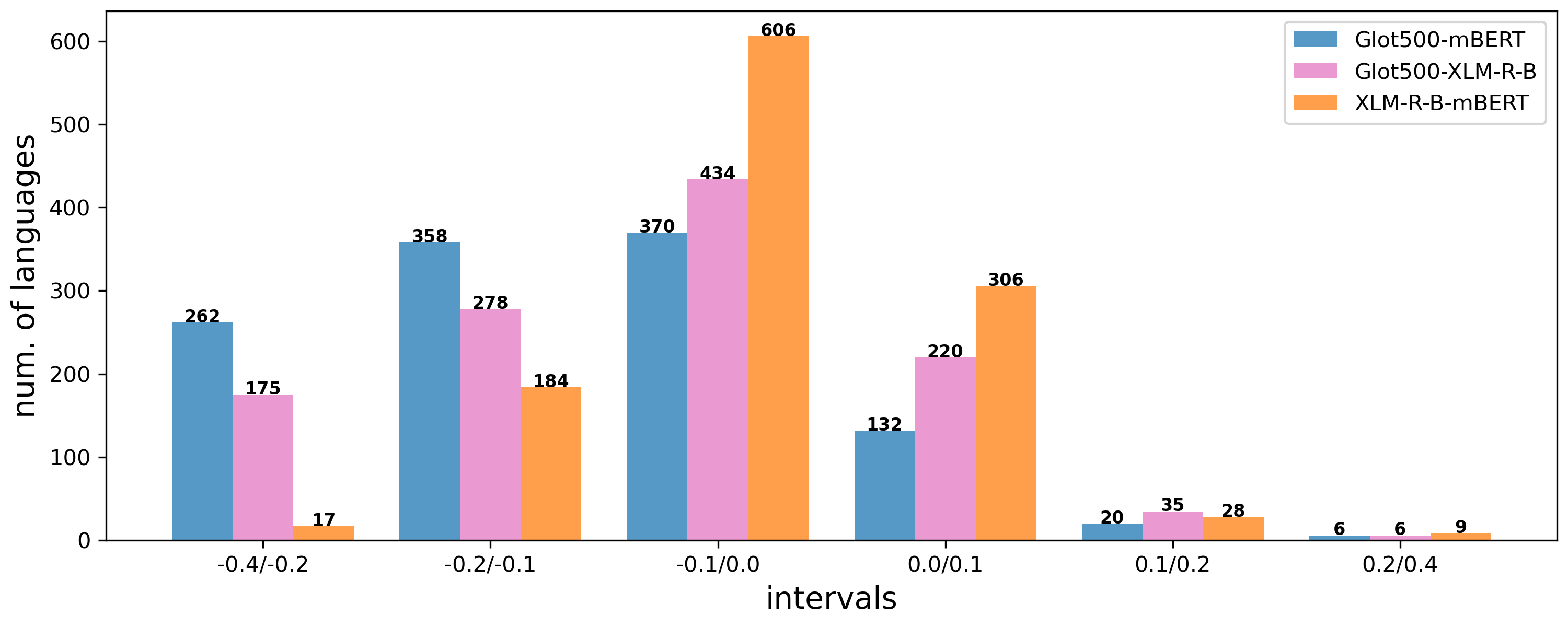}
\caption{F1-differences of in-language learning for 1504 languages. We split the F1-differences between two models into six intervals: -0.4/-0.2, -0.2/-0.1, -0.1/0.0, 0.0/0.1,0.1/0.2, 0.2/0.4. Each bar represents the comparison between a pair of models.}
\label{fig:in_language_four_models}
\end{figure}

Figure \ref{fig:in_language_four_models} shows differences in F1
for the four models on head,
\tailone, and \tailtwo languages. We see that Glot500 and
XLM-R-Base have better performance than mBERT on head languages
(most differences are positive). XLM-R-Base outperforms
Glot500 slightly on 11 head languages. On \tailone
languages, Glot500m outperforms the other three models as
expected with a larger number of positive differences. On
\tailtwo languages, mBERT has better performance than the
other three models. This may be due to the other models having larger numbers of parameters and thus being more prone to overfitting.


\enote{hs}{figure 4: please provide subheadings for the
  three subfgiures: head languages, glot500-only languages,
  tail languages}

\subsubsection{Influence of training set size}
To investigate  the influence of the training set size,
we conduct in-language experiments with 20
selected languages, 10 head and 10 \tailtwo
languages. We select languages based on: 1) their inclusion in pretrained multilingual models, specifically whether they are pretrained by the four mentioned SOTA PMLMs or not. 2) variation in typology to ensure coverage of different language types and families, including a) high resource languages, and 2) low resource languages pretrained by the SOTA PMLMs, as well as 3) low resource languages and 4) resource-scarce languages not covered by the SOTA PMLMs like Hixkaryana. We present the iso codes, writing systems and language families of the 20 languages in table \ref{tab:lan_family_script}. The languages are selected to represent
11 different writing systems (Latin, Chinese, Korean, Japanese, Basque,
Hebrew, Arabic, Malayalam, Cyrillic, Devanagari and Burmese) and 13
language families (Indo-European, Sino-Tibetan, Koreanic,
Japanic, Basque, Dravidian, Iroquoian, Turkic, Cariban,
Uralic, Austronesian, Uto-Aztecan and Central Sudanic). Tables \ref{tab:in_language_mbert} and
\ref{tab:in_language_xlmr_b} show the results of zero-shot
transfer and in-language experiments using mBERT and XLM-R-Base for the
selected languages. As expected, the in-language
performance improves when the training set becomes
larger. Interestingly, zero-shot transfer performance of head
languages is comparable to in-language setting with 100
samples for mBert and with 400 samples for XLM-R-Base,
which indicates that
models with more parameters may require more in-language data to
reach a comparable level with zero-shot transfer performance.
In addition, observing the average
zero-shot transfer performance of mBert and XLM-R-Base,
XLM-R-Base achieves higher scores on both head
and \tailtwo languages, this might indicate a
better overall performance of XLM-R-Base on Taxi1500 classification
task. Moreover, the zero-shot transfer results on both
models show that head languages consistently outperform
\tailtwo languages, which reflects both models'
better generalization capability on languages in their pretraining data.

\begin{table*}[h!]
  \centering
  \resizebox{\textwidth}{!}{
  \small
  \begin{tabular}{l|l|l|l||l|l|l|l}
    head lang. & iso  & Script  & Family & \tailtwo lang. & iso  & Script & Family  \\
    \midrule
German    & deu & Latin     & Indo-European & Cherokee         & chr & Cherokee   & Iroquoian        \\
Basque    & eus & Latin     & Basque        & Gagauz           & gag & Latin      & Turkic           \\
Hebrew    & heb & Hebrew    & Afro-Asiatic  & Hixkaryana       & hix & Latin      & Cariban          \\
Japanese  & jpn & Japanese  & Japanic       & Nga La           & hlt & Latin      & Sino-Tibetan     \\
Kazakh    & kaz & Cyrilic   & Turkic        & Komi-Zyrian      & kpv & Cyrilic    & Uralic           \\
Korean    & kor & Korean    & Koreanic      & Kumyk            & kum & Cyrilic    & Turkic           \\
Malayalam & mal & Malayalam & Dravidian     & Aringa           & luc & Latin      & Central Sudanic  \\
Burmese   & mya & Burmese   & Indo-European & Magahi           & mag & Devanagari & Indo-European    \\
Persian   & pes & Arabic    & Indo-European & Dibabawon Manobo & mbd & Latin      & Austronesian     \\
Chinese   & zho & Chinese   & Sino-Tebietan & Middle Watut     & npl & Latin      & Uto-Aztecan     
    \end{tabular}
}
\caption{\label{tab:lan_family_script}
An overview of selected 20 languages from 11 different writing systems and 13 language families
}
\end{table*}

\begin{table*}[h!]
  \centering
  \resizebox{\textwidth}{!}{
  \small
  \begin{tabular}{m{5mm}|m{10mm}|m{5mm}m{5mm}m{5mm}m{5mm}m{5mm}m{5mm} || m{5mm}|m{10mm}|m{5mm}m{5mm}m{5mm}m{5mm}m{5mm}m{5mm}}
            head& transfer & \multicolumn{6}{c}{in-language training}     & \tailtwo& transfer &\multicolumn{6}{c}{in-language training} \\
    lang.   &     & 50   & 100  & 200  & 400  & 600  & 860 & lang.   &     & 50   & 100  & 200  & 400  & 600  & 860  \\
    \midrule
    deu & 0.39 & 0.20 & 0.13 & 0.34 & 0.42 & 0.44 & 0.52 & chr & 0.05 & 0.24 & 0.21 & 0.29 & 0.35 & 0.30 & 0.35 \\
    eus & 0.17 & 0.15 & 0.12 & 0.31 & 0.44 & 0.46 & 0.43 & gag & 0.12 & 0.21 & 0.29 & 0.35 & 0.39 & 0.45 & 0.38 \\
    heb & 0.36 & 0.24 & 0.24 & 0.36 & 0.33 & 0.38 & 0.41 & hix & 0.07 & 0.30 & 0.27 & 0.35 & 0.35 & 0.39 & 0.41 \\
    jpn & 0.39 & 0.37 & 0.40 & 0.32 & 0.49 & 0.63 & 0.66 & hlt & 0.08 & 0.16 & 0.25 & 0.33 & 0.34 & 0.44 & 0.49 \\
    kaz & 0.29 & 0.30 & 0.36 & 0.38 & 0.50 & 0.48 & 0.48 & kpv & 0.08 & 0.19 & 0.24 & 0.45 & 0.41 & 0.39 & 0.46 \\
    kor & 0.41 & 0.36 & 0.36 & 0.45 & 0.56 & 0.50 & 0.60 & kum & 0.14 & 0.28 & 0.27 & 0.35 & 0.37 & 0.42 & 0.46 \\
    mal & 0.09 & 0.13 & 0.25 & 0.25 & 0.31 & 0.35 & 0.34 & luc & 0.08 & 0.27 & 0.23 & 0.46 & 0.41 & 0.45 & 0.35 \\
    mya & 0.22 & 0.32 & 0.31 & 0.41 & 0.41 & 0.40 & 0.46 & mag & 0.19 & 0.14 & 0.38 & 0.38 & 0.37 & 0.43 & 0.34 \\
    pes & 0.43 & 0.30 & 0.36 & 0.55 & 0.53 & 0.52 & 0.56 & mbd & 0.08 & 0.18 & 0.33 & 0.36 & 0.36 & 0.39 & 0.42 \\
    zho & 0.36 & 0.24 & 0.46 & 0.47 & 0.62 & 0.54 & 0.59 & npl & 0.06 & 0.21 & 0.30 & 0.38 & 0.39 & 0.40 & 0.40 \\
    \midrule
    \textbf{avg.} & 0.31 & 0.26 & 0.30 & 0.38 & 0.46 & 0.47 & 0.51 & \textbf{avg.} & 0.10 & 0.22 & 0.28 & 0.37 & 0.37 & 0.41 & 0.41
    \end{tabular}
}
\caption{\label{tab:in_language_mbert}
Results of zero-shot transfer and in-language fine-tuning experiments using mBERT for 20 selected languages, 10 head (left): German, Basque, Hebrew, Japanese, Kazakh, Korean, Malayalam, Burmese, Persian and Chinese, and 10 \tailtwo (right): Cherokee, Gagauz, Hixkaryana, Nga La,	Komi-Zyrian, Kumyk, Aringa, Magahi, Dibabawon Manobo and Middle Watut. The numbers in the table header indicate the size of target language training data: 860 means the full training set.
}
\end{table*}

\begin{table*}[h!]
  \centering
  \resizebox{\textwidth}{!}{
  \small
  \begin{tabular}{m{5mm}|m{10mm}|m{5mm}m{5mm}m{5mm}m{5mm}m{5mm}m{5mm} || m{5mm}|m{10mm}|m{5mm}m{5mm}m{5mm}m{5mm}m{5mm}m{5mm}}
            head& transfer & \multicolumn{6}{c}{in-language training}     & \tailtwo& transfer &\multicolumn{6}{c}{in-language training} \\
    lang.   &     & 50   & 100  & 200  & 400  & 600  & 860 & lang.   &     & 50   & 100  & 200  & 400  & 600  & 860  \\
    \midrule
    deu & 0.52 & 0.16 & 0.18 & 0.43 & 0.49 & 0.52 & 0.51 & chr & 0.09 & 0.15 & 0.20 & 0.15 & 0.24 & 0.21 & 0.28 \\
    eus & 0.26 & 0.09 & 0.26 & 0.25 & 0.34 & 0.37 & 0.34 & gag & 0.33 & 0.17 & 0.13 & 0.14 & 0.45 & 0.32 & 0.54 \\
    heb & 0.15 & 0.10 & 0.13 & 0.18 & 0.16 & 0.33 & 0.35 & hix & 0.06 & 0.18 & 0.17 & 0.22 & 0.3 & 0.43 & 0.49 \\
    jpn & 0.62 & 0.25 & 0.39 & 0.53 & 0.57 & 0.61 & 0.68 & hlt & 0.05 & 0.14 & 0.07 & 0.19 & 0.40 & 0.20 & 0.50 \\
    kaz & 0.57 & 0.23 & 0.35 & 0.47 & 0.41 & 0.55 & 0.56 & kpv & 0.09 & 0.09 & 0.21 & 0.23 & 0.41 & 0.38 & 0.53 \\
    kor & 0.63 & 0.35 & 0.55 & 0.58 & 0.65 & 0.53 & 0.70 & kum & 0.13 & 0.13 & 0.17 & 0.22 & 0.27 & 0.37 & 0.45 \\
    mal & 0.07 & 0.10 & 0.13 & 0.22 & 0.08 & 0.21 & 0.24 & luc & 0.11 & 0.12 & 0.11 & 0.30 & 0.30 & 0.39 & 0.39 \\
    mya & 0.42 & 0.18 & 0.30 & 0.21 & 0.45 & 0.45 & 0.64 & mag & 0.38 & 0.11 & 0.23 & 0.41 & 0.48 & 0.38 & 0.51 \\
    pes & 0.66 & 0.17 & 0.55 & 0.47 & 0.65 & 0.64 & 0.71 & mbd & 0.11 & 0.18 & 0.14 & 0.25 & 0.30 & 0.30 & 0.38 \\
    zho & 0.63 & 0.33 & 0.49 & 0.52 & 0.45 & 0.51 & 0.68 & npl & 0.05 & 0.14 & 0.08 & 0.25 & 0.41 & 0.41 & 0.43 \\
    \midrule
    \textbf{avg.} & 0.45 & 0.20 & 0.33 & 0.39 & 0.43 & 0.47 & 0.54 & \textbf{avg.} & 0.14 & 0.14 & 0.15 & 0.24 & 0.36 & 0.34 & 0.45
    \end{tabular}
}
\caption{\label{tab:in_language_xlmr_b}
Results of zero-shot transfer and in-language fine-tuning experiments using XLM-R-Base for 20 selected languages, 10 head (left): German, Basque, Hebrew, Japanese, Kazakh, Korean, Malayalam, Burmese, Persian and Chinese, and 10 \tailtwo (right): Cherokee, Gagauz, Hixkaryana, Nga La,	Komi-Zyrian, Kumyk, Aringa, Magahi, Dibabawon Manobo and Middle Watut. The numbers in the table header indicate the size of target language training data: 860 means the full training set.
}
\end{table*}

\vspace{0.5cm}

\subsubsection{Analysis by Language Family}
\begin{sidewaysfigure*}
  \centering
  \includegraphics[width=0.8\columnwidth]{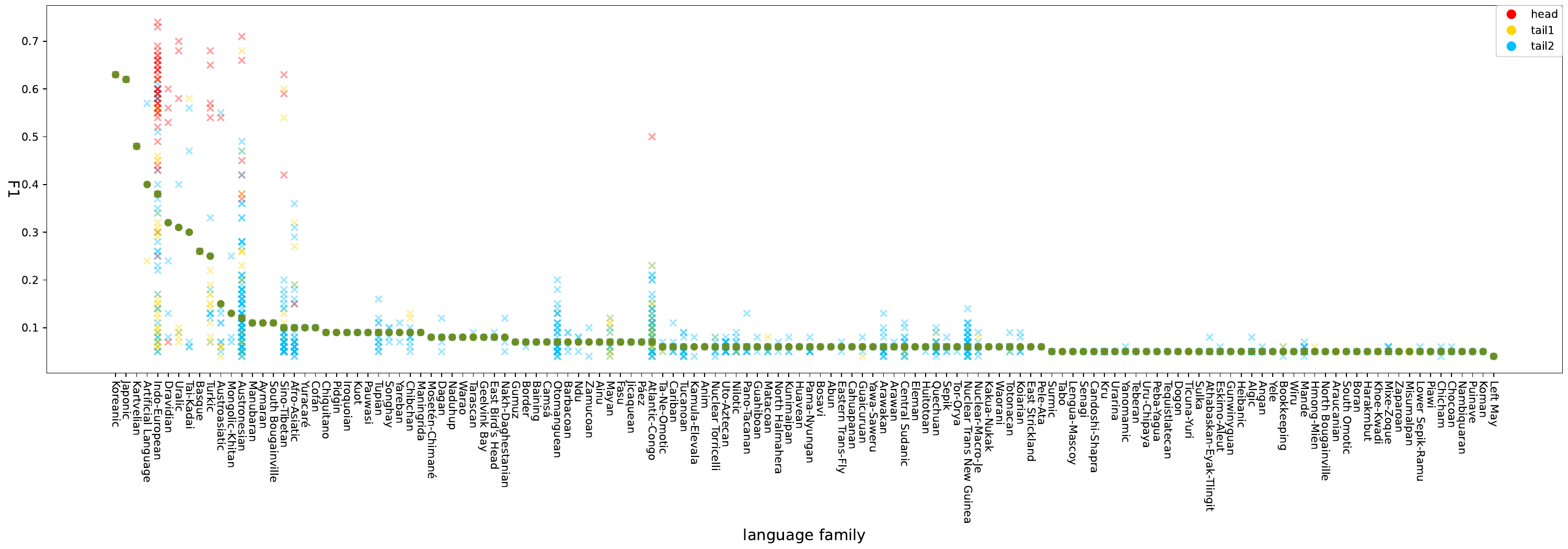}
  \includegraphics[width=0.8\columnwidth]{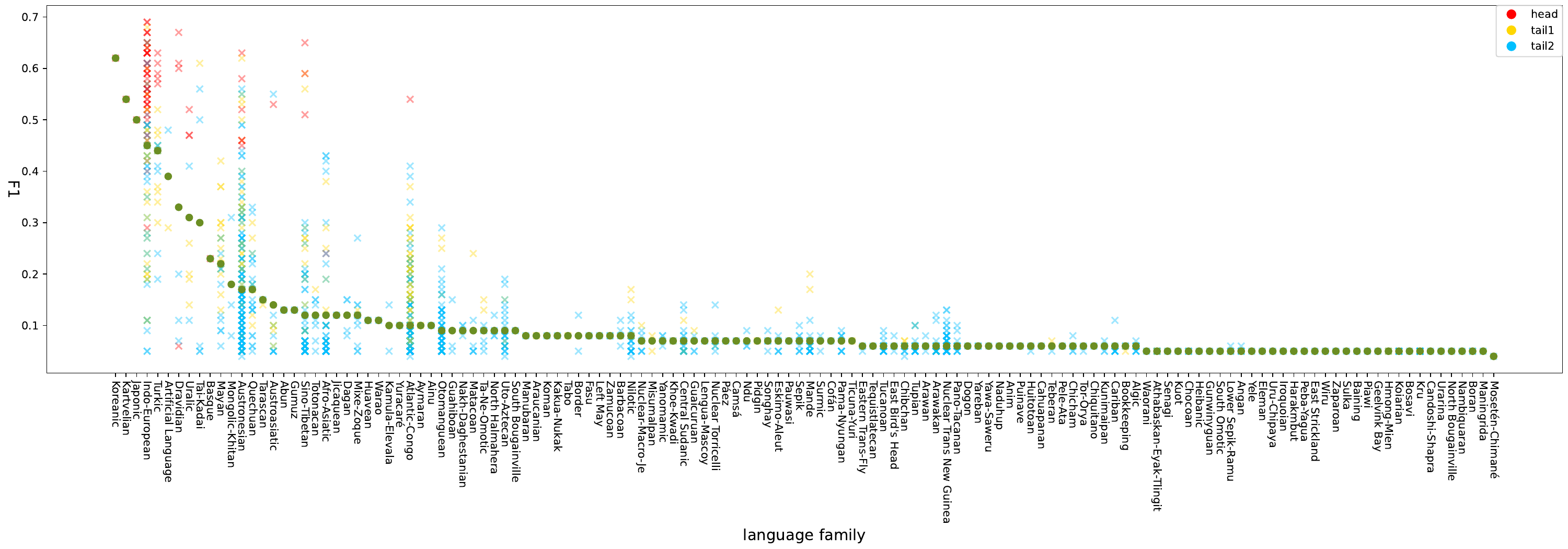}
  \caption{Zero shot transfer learning: F1 of XLM-R-Base (top) and Glot500 (bottom). Each small dot represents a
  language, each large dot an average per family. Families
  are sorted by
  F1. Red, yellow and blue represent head, \tailone and \tailtwo languages respectively.}
  \label{fig:zeroshot_xlmr_base_Glot500}
\end{sidewaysfigure*}

\begin{sidewaysfigure*}
  \centering
  \includegraphics[width=0.8\columnwidth]{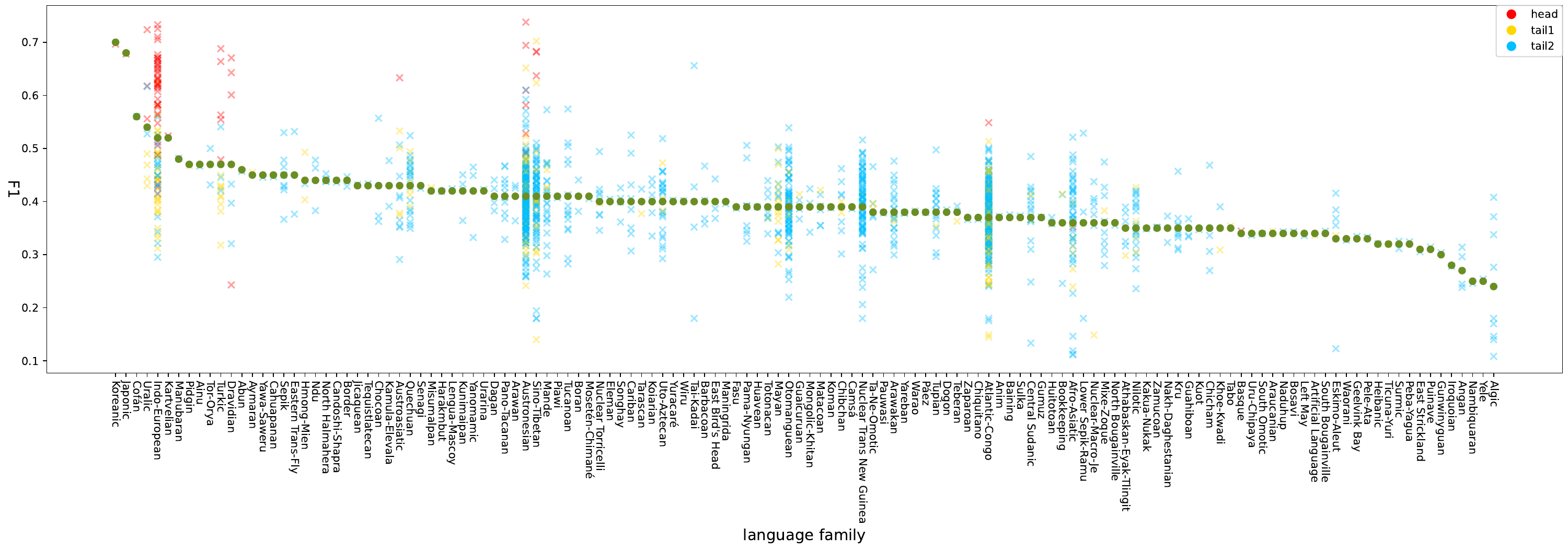}
  \includegraphics[width=0.8\columnwidth]{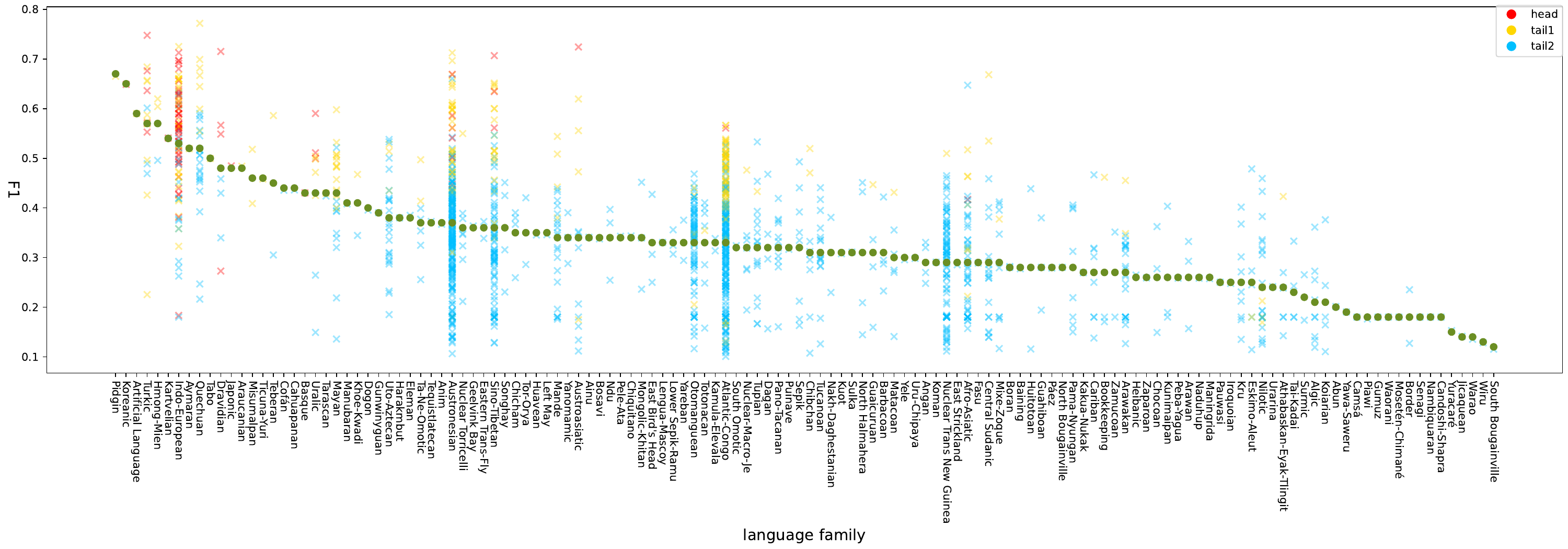}
  \caption{In-language results: F1 of XLM-R-Base (top) and Glot500 (bottom). Each small dot represents a
  language, each large dot an average per family. Families
  are sorted by F1. Red, yellow and blue represent head, \tailone and \tailtwo languages respectively.}
  \label{fig:in_language_xlmr_base_Glot500}
\end{sidewaysfigure*}

In Figures \ref{fig:zeroshot_xlmr_base_Glot500} and
\ref{fig:in_language_xlmr_base_Glot500}, we present
zero-shot transfer and in-language results of all languages based on
their families \citep{hammarstrom2015glottolog} on XLM-R-Base and Glot500.
For almost all families, the performance on head languages is significantly higher than that of \tailone and \tailtwo
languages. The Indo-European family outperforms other
language families not only on head languages but also on
\tailone and \tailtwo languages. We suppose the reason is
that the four evaluated models are pre-trained with more
Indo-European languages, which increases the performance of
this family. We also notice that XLM-R-Large tends to
perform worse than the other three models on most
languages. We think this could be due to its larger number of parameters, which makes it prone to overfitting on our small dataset.
Interestingly, by comparing zero-shot transfer and in-language results of XLM-R-Base, we find that
languages that are extremely low-resource and use non-Latin scripts (e.g. Yawa-Saweru, Lengua-Mascoy, and Hmong-Mien) have
significant performance increases (around 0.4)
when they are trained with in-language data. This indicates
that the four models do not perform as well on non-Latin scripts as on Latin scripts.

\enote{hs}{the captions of zeroshot\_xlmr\_base\_Glot500 and
  in\_language\_xlmr\_base\_Glot500 are confusing with respect
  to F1. Do you mean F1 or macro F1 when you say ``F1''? Do
  you only show macro F1 or do you also show F1?}
\enote{cl}{I changed macro-F1 to F1, because all results in this paper are macro-F1}



\section{Conclusion}
A bottleneck for the evaluation of multilingual models is the lack of evaluation data for many low-resource languages.
Since annotating data for every language is a prohibitively
expensive and an unrealistic approach, there is an increasing interest in evaluation data for low-resource languages.
In this paper, we introduce a text classification dataset, Taxi1500, which consists of annotated Bible verses in 1504 languages. We obtain labels for the English verses through crowdsourcing and project the labels to all other languages, making use of parallel verses.
We present several use cases of Taxi1500 by conducting a thorough evaluation of four multilingual language models.
We hope the high language coverage of Taxi1500 will encourage research on multilingual language models, and especially, benefit low-resource languages that are, till now, neglected due to the lack of evaluation data. 


\section{Limitations}
While the high degree of parallelism in the PBC makes it a 
valuable tool for massively multilingual applications, 
such as the building of Taxi1500, it is not perfect. 
One limitation is the religious domain of the Bible, which means keywords specific to the domain may be exploited.
Also, we are restricted to the New Testament, as many languages do not have a translated Old Testament in the PBC.
Given that some extremely low-resource languages do not have complete translations, the actual number of available verses varies for each language. 
However, since the Bible is arguably the most translated book in the world, we regard it as a suitable resource for an initiative to build highly parallel dataset like ours.

\section{Failure analysis}
In this section, we present a path of methods used when designing categories for the sentence classification task and the difficulties met during the data annotation process.

\subsection{Category design}
We have developed seven versions of topics in total (shown in table \ref{tab:category_version}), each new one based on the refinements of the previous version. This is done by collecting feedback from NLP experts and workers from Amazon MTurk.

We conclude a few of the reasons why earlier versions of topics fail as follows:
\begin{enumerate}
    \item Lack of domain knowledge. In the first version, we read the Bible and come up with our own topics. Due to limited background knowledge of the religious domain, the first topics are rather arbitrary and do not cover sufficient verses. For the second version, we consult a theologist and online preaching websites that contain a large number of topics, from which we select ones that cover a high number of verses.
    \item Obscure or abstract topics. Verses in v2 are collected from an online preaching website, ProPreacher\footnote{\url{https://www.propreacher.com/100}}. We sample 100 verses and ask for feedback from crowdsourcing workers on them. Many workers think several topics are very hard to understand or recognize from the verses, for example, \textit{Eschatology, Philosophy, Theology} and \textit{Moral}. Therefore, v3 deletes four abstract topics, \textit{Eschatology, Philosophy, Theology}, and \textit{Moral}, and adds five new ones, \textit{Repentance, Friendship, Thankfulness, Forgiveness} and \textit{Suffering}, which are easier to understand.
    \item Overlap between topics. v4 is the version we use to crowdsource annotation on Amazon Mechanical Turk. However, we get feedback indicating that many verses can be assigned multiple topics, such as \textit{Violence} and \textit{Conflict}. Therefore, in v5 and v6 we combine similar topics in v4 and change the names of several labels. v7 is the version that contains the final topics in Taxi1500.
\end{enumerate}

\begin{table}[h!]
\resizebox{\textwidth}{!}{
\begin{tabular}{m{0.1\textwidth}m{0.6\textwidth}m{0.2\textwidth}}
 \hline
 Version & Topics & Num. topics\\ [0.5ex] 
 \hline
 v1 & Rules, Phenomenon, Conflict, Relation, Place, Character, Reward, Punishment, Command & 9\\
 \hline
 v2 & Eschatology, Grace, Family, Creation, Philosophy, Revival, Cults, Compromise, Persecution, Hospitality, Conflicts, Theology, Morals, Commandments, Sacrifice & 15\\ 
 \hline
 v3 & Creation, Grace, Violence, Conflict, Hospitality, Sacrifice, Heresy, Repentance, Faith, Suffering, Forgiveness, Thankfulness, Friendship, Temptation & 14\\
 \hline
 v4 & Creation, Grace, Violence, Conflict, Hospitality, Sacrifice, Heresy, Repentance, Faith, Suffering, Forgiveness, Thankfulness & 12\\
 \hline
 v5 & Creation, Commandment, Genealogy, Violence, Sacrifice, Money, Salvation, Sin & 8\\
 \hline
 v6 & Creation, Commandment, Genealogy, Violence, Sacrifice, Money,  Grace, Sin & 8\\ 
 \hline
 v7 & Recommendation, Faith, Description, Sin, Grace, Violence & 6\\ 
 \hline
\end{tabular}
 }
\caption{An overview of different versions of designed categories. v7 is the final version for Taxi1500}
\label{tab:category_version}
\end{table}

\noindent
\subsection{Data annotation}
We choose Amazon Mechanical Turk (MTurk) for data annotation because of the availability of a large number of native English speakers. Besides, its usage is well documented in online tutorials for building annotation projects.

Based on our experience, we provide some tips based on our failure when using Amazon MTurk as follows:
\begin{enumerate}
    \item Although the lowest payment for every HIT is 0.01 US dollars, workers seldom do the task for the minimum payment. It is recommended to set a higher payment if possible.
    \item It is not advisable to reject HITs if the requester is new to MTurk, lest the requester's approval rate drops significantly, which will attract fewer workers.
    \item Clear instruction and a qualification test prior to permitting workers to annotate are strongly recommended for high-quality data.
    \item It is better to test with a smaller batch first before uploading all data for annotation because there can be errors in the instruction or the data submitted.
    \item Workers may have valuable opinions about the task and it is a good idea to contact them for feedback.
\end{enumerate}

\section{Ethics Statement}
In this work, we introduce a new multilingual text classification dataset based on the Parallel Bible Corpus. The data is partially annotated by workers from the Amazon mTurk platform, who are rewarded fairly for their work (\$0.2 per sentence). Our dataset contains Bible verses for which we estimate a low risk of tracing to specific individuals and are intended exclusively for the evaluation of NLP tasks concerning the supported languages. We therefore do not expect any ethical issues with our dataset.

\section*{Declarations}
\bmhead{Funding}
This work is funded by the European Research Council (grant no. 740516).

\bmhead{Conflicts of interest}
We do not foresee any conflicts of interest with this work.

\bmhead{Author contributions}
Main manuscript text: Chunlan Ma;
Tables \ref{table: multilingual benchamrk},
 \ref{tab:definition_verses},
\ref{fig:confusion matrix},
\ref{tab:language_verses},
\ref{tab:in_language_mbert},
\ref{tab:in_language_xlmr_b}: Chunlan Ma; Tables \ref{tab:k_alpha}, \ref{tab:example_verses}, \ref{tab:lan_family_script}: Haotian Ye; Figures \ref{fig:zeroshot_four_models}, \ref{fig:in_language_four_models}: Haotian Ye; Review and revision of the manuscript: all authors.

\bmhead{Data availability}
Part of the data analyzed for the current study is not publicly available due to copyright restrictions. It is available from the corresponding author upon reasonable request.




\bibliography{sn-bibliography, bibliography}

\begin{thebibliography}{35}
\providecommand{\natexlab}[1]{#1}
\providecommand{\url}[1]{{#1}}
\providecommand{\urlprefix}{URL }
\providecommand{\doi}[1]{\url{https://doi.org/#1}}
\providecommand{\eprint}[2][]{\url{#2}}
 \bibcommenthead

\bibitem[{Adebara et~al(2022)Adebara, Elmadany, Abdul-Mageed, and
  Inciarte}]{adebara2022serengeti}
Adebara I, Elmadany A, Abdul-Mageed M, et~al (2022) Serengeti: Massively
  multilingual language models for africa. \eprint{2212.10785}

\bibitem[{Agi{\'c} and Vuli{\'c}(2019)}]{agic-vulic-2019-jw300}
Agi{\'c} {\v{Z}}, Vuli{\'c} I (2019) {JW}300: A wide-coverage parallel corpus
  for low-resource languages. In: Proceedings of the 57th Annual Meeting of the
  Association for Computational Linguistics. Association for Computational
  Linguistics, Florence, Italy, pp 3204--3210, \doi{10.18653/v1/P19-1310},
  \urlprefix\url{https://aclanthology.org/P19-1310}

\bibitem[{Alabi et~al(2022)Alabi, Adelani, Mosbach, and
  Klakow}]{alabi-etal-2022-adapting}
Alabi JO, Adelani DI, Mosbach M, et~al (2022) Adapting pre-trained language
  models to {A}frican languages via multilingual adaptive fine-tuning. In:
  Proceedings of the 29th International Conference on Computational
  Linguistics. International Committee on Computational Linguistics, Gyeongju,
  Republic of Korea, pp 4336--4349,
  \urlprefix\url{https://aclanthology.org/2022.coling-1.382}

\bibitem[{Artetxe and Schwenk(2019)}]{artetxe-schwenk-2019-massively}
Artetxe M, Schwenk H (2019) Massively multilingual sentence embeddings for
  zero-shot cross-lingual transfer and beyond. Transactions of the Association
  for Computational Linguistics 7:597--610. \doi{10.1162/tacl_a_00288},
  \urlprefix\url{https://aclanthology.org/Q19-1038}

\bibitem[{Artetxe et~al(2019)Artetxe, Ruder, and Yogatama}]{artetxe2019cross}
Artetxe M, Ruder S, Yogatama D (2019) On the cross-lingual transferability of
  monolingual representations. arXiv preprint arXiv:191011856

\bibitem[{Clark et~al(2020)Clark, Luong, Le, and Manning}]{clark2020electra}
Clark K, Luong MT, Le QV, et~al (2020) Electra: Pre-training text encoders as
  discriminators rather than generators. \eprint{2003.10555}

\bibitem[{Conneau et~al(2020)Conneau, Khandelwal, Goyal, Chaudhary, Wenzek,
  Guzm{\'a}n, Grave, Ott, Zettlemoyer, and
  Stoyanov}]{conneau-etal-2020-unsupervised}
Conneau A, Khandelwal K, Goyal N, et~al (2020) Unsupervised cross-lingual
  representation learning at scale. In: Proceedings of the 58th Annual Meeting
  of the Association for Computational Linguistics. Association for
  Computational Linguistics, Online, pp 8440--8451,
  \doi{10.18653/v1/2020.acl-main.747},
  \urlprefix\url{https://aclanthology.org/2020.acl-main.747}

\bibitem[{De~Marneffe et~al(2014)De~Marneffe, Dozat, Silveira, Haverinen,
  Ginter, Nivre, and Manning}]{de2014universal}
De~Marneffe MC, Dozat T, Silveira N, et~al (2014) Universal stanford
  dependencies: A cross-linguistic typology. In: Proceedings of the Ninth
  International Conference on Language Resources and Evaluation (LREC'14), pp
  4585--4592

\bibitem[{De~Marneffe et~al(2021)De~Marneffe, Manning, Nivre, and
  Zeman}]{de2021universal}
De~Marneffe MC, Manning CD, Nivre J, et~al (2021) Universal dependencies.
  Computational linguistics 47(2):255--308

\bibitem[{Devlin et~al(2018)Devlin, Chang, Lee, and
  Toutanova}]{DBLP:journals/corr/abs-1810-04805}
Devlin J, Chang M, Lee K, et~al (2018) {BERT:} pre-training of deep
  bidirectional transformers for language understanding. CoRR abs/1810.04805.
  \urlprefix\url{http://arxiv.org/abs/1810.04805},
  {\href{https://arxiv.org/abs/1810.04805}{{https://arxiv.org/abs/1810.04805}}}

\bibitem[{Eisenschlos et~al(2019)Eisenschlos, Ruder, Czapla, Kardas, Gugger,
  and Howard}]{DBLP:journals/corr/abs-1909-04761}
Eisenschlos J, Ruder S, Czapla P, et~al (2019) Multifit: Efficient
  multi-lingual language model fine-tuning. CoRR abs/1909.04761.
  \urlprefix\url{http://arxiv.org/abs/1909.04761},
  {\href{https://arxiv.org/abs/1909.04761}{{https://arxiv.org/abs/1909.04761}}}

\bibitem[{Hammarstr{\"o}m(2015)}]{hammarstrom2015glottolog}
Hammarstr{\"o}m H (2015) Glottolog: A free, online, comprehensive bibliography
  of the world's languages. In: 3rd International Conference on Linguistic and
  Cultural Diversity in Cyberspace, UNESCO, pp 183--188

\bibitem[{Hu et~al(2020)Hu, Ruder, Siddhant, Neubig, Firat, and
  Johnson}]{DBLP:journals/corr/abs-2003-11080}
Hu J, Ruder S, Siddhant A, et~al (2020) {XTREME:} {A} massively multilingual
  multi-task benchmark for evaluating cross-lingual generalization. CoRR
  abs/2003.11080. \urlprefix\url{https://arxiv.org/abs/2003.11080},
  {\href{https://arxiv.org/abs/2003.11080}{{https://arxiv.org/abs/2003.11080}}}

\bibitem[{ImaniGooghari et~al(2023)ImaniGooghari, Lin, Kargaran, Severini,
  Sabet, Kassner, Ma, Schmid, Martins, Yvon, and
  Sch{\"u}tze}]{imani-etal-2023-glot500}
ImaniGooghari A, Lin P, Kargaran AH, et~al (2023) Glot500: Scaling multilingual
  corpora and language models to 500 languages. In: Proceedings of the 61st
  Annual Meeting of the Association for Computational Linguistics.

\bibitem[{Joshi et~al(2020)Joshi, Santy, Budhiraja, Bali, and
  Choudhury}]{joshi-etal-2020-state}
Joshi P, Santy S, Budhiraja A, et~al (2020) The state and fate of linguistic
  diversity and inclusion in the {NLP} world. In: Proceedings of the 58th
  Annual Meeting of the Association for Computational Linguistics. Association
  for Computational Linguistics, Online, pp 6282--6293,
  \doi{10.18653/v1/2020.acl-main.560},
  \urlprefix\url{https://aclanthology.org/2020.acl-main.560}

\bibitem[{Klementiev et~al(2012)Klementiev, Titov, and
  Bhattarai}]{klementiev2012inducing}
Klementiev A, Titov I, Bhattarai B (2012) Inducing crosslingual distributed
  representations of words. In: Proceedings of COLING 2012, pp 1459--1474

\bibitem[{Koehn(2005)}]{koehn-2005-europarl}
Koehn P (2005) {E}uroparl: A parallel corpus for statistical machine
  translation. In: Proceedings of Machine Translation Summit X: Papers, Phuket,
  Thailand, pp 79--86,
  \urlprefix\url{https://aclanthology.org/2005.mtsummit-papers.11}

\bibitem[{Kunchukuttan et~al(2018)Kunchukuttan, Mehta, and
  Bhattacharyya}]{kunchukuttan-etal-2018-iit}
Kunchukuttan A, Mehta P, Bhattacharyya P (2018) The {IIT} {B}ombay
  {E}nglish-{H}indi parallel corpus. In: Proceedings of the Eleventh
  International Conference on Language Resources and Evaluation ({LREC} 2018).
  European Language Resources Association (ELRA), Miyazaki, Japan,
  \urlprefix\url{https://aclanthology.org/L18-1548}

\bibitem[{Lample and Conneau(2019)}]{DBLP:journals/corr/abs-1901-07291}
Lample G, Conneau A (2019) Cross-lingual language model pretraining. CoRR
  abs/1901.07291. \urlprefix\url{http://arxiv.org/abs/1901.07291},
  {\href{https://arxiv.org/abs/1901.07291}{{https://arxiv.org/abs/1901.07291}}}

\bibitem[{Lewis et~al(2004)Lewis, Yang, Russell-Rose, and Li}]{lewis2004rcv1}
Lewis DD, Yang Y, Russell-Rose T, et~al (2004) Rcv1: A new benchmark collection
  for text categorization research. Journal of machine learning research
  5(Apr):361--397

\bibitem[{Lewis et~al(2020)Lewis, Oguz, Rinott, Riedel, and
  Schwenk}]{lewis-etal-2020-mlqa}
Lewis P, Oguz B, Rinott R, et~al (2020) {MLQA}: Evaluating cross-lingual
  extractive question answering. In: Proceedings of the 58th Annual Meeting of
  the Association for Computational Linguistics. Association for Computational
  Linguistics, Online, pp 7315--7330, \doi{10.18653/v1/2020.acl-main.653},
  \urlprefix\url{https://aclanthology.org/2020.acl-main.653}

\bibitem[{Liu et~al(2019)Liu, Ott, Goyal, Du, Joshi, Chen, Levy, Lewis,
  Zettlemoyer, and Stoyanov}]{DBLP:journals/corr/abs-1907-11692}
Liu Y, Ott M, Goyal N, et~al (2019) Roberta: {A} robustly optimized {BERT}
  pretraining approach. CoRR abs/1907.11692.
  \urlprefix\url{http://arxiv.org/abs/1907.11692},
  {\href{https://arxiv.org/abs/1907.11692}{{https://arxiv.org/abs/1907.11692}}}

\bibitem[{Mayer and Cysouw(2014)}]{mayer-cysouw-2014-creating}
Mayer T, Cysouw M (2014) Creating a massively parallel {B}ible corpus. In:
  Proceedings of the Ninth International Conference on Language Resources and
  Evaluation ({LREC}'14). European Language Resources Association (ELRA),
  Reykjavik, Iceland, pp 3158--3163,
  \urlprefix\url{http://www.lrec-conf.org/proceedings/lrec2014/pdf/220_Paper.pdf}

\bibitem[{Mogadala and Rettinger(2016)}]{mogadala2016bilingual}
Mogadala A, Rettinger A (2016) Bilingual word embeddings from parallel and
  non-parallel corpora for cross-language text classification. In: Proceedings
  of the 2016 Conference of the North American Chapter of the Association for
  Computational Linguistics: Human Language Technologies, pp 692--702

\bibitem[{Nzeyimana and Rubungo(2022)}]{Nzeyimana_2022}
Nzeyimana A, Rubungo AN (2022) {KinyaBERT}: a morphology-aware kinyarwanda
  language model. In: Proceedings of the 60th Annual Meeting of the Association
  for Computational Linguistics (Volume 1: Long Papers). Association for
  Computational Linguistics, \doi{10.18653/v1/2022.acl-long.367},
  \urlprefix\url{https://doi.org/10.18653%2Fv1%2F2022.acl-long.367}

\bibitem[{Ogueji et~al(2021)Ogueji, Zhu, and Lin}]{ogueji-etal-2021-small}
Ogueji K, Zhu Y, Lin J (2021) Small data? no problem! exploring the viability
  of pretrained multilingual language models for low-resourced languages. In:
  Proceedings of the 1st Workshop on Multilingual Representation Learning.
  Association for Computational Linguistics, Punta Cana, Dominican Republic, pp
  116--126, \doi{10.18653/v1/2021.mrl-1.11},
  \urlprefix\url{https://aclanthology.org/2021.mrl-1.11}

\bibitem[{Pan et~al(2017)Pan, Zhang, May, Nothman, Knight, and
  Ji}]{pan-etal-2017-cross}
Pan X, Zhang B, May J, et~al (2017) Cross-lingual name tagging and linking for
  282 languages. In: Proceedings of the 55th Annual Meeting of the Association
  for Computational Linguistics (Volume 1: Long Papers). Association for
  Computational Linguistics, Vancouver, Canada, pp 1946--1958,
  \doi{10.18653/v1/P17-1178}, \urlprefix\url{https://aclanthology.org/P17-1178}

\bibitem[{Petrov et~al(2012)Petrov, Das, and
  McDonald}]{petrov-etal-2012-universal}
Petrov S, Das D, McDonald R (2012) A universal part-of-speech tagset. In:
  Proceedings of the Eighth International Conference on Language Resources and
  Evaluation ({LREC}'12). European Language Resources Association (ELRA),
  Istanbul, Turkey, pp 2089--2096,
  \urlprefix\url{http://www.lrec-conf.org/proceedings/lrec2012/pdf/274_Paper.pdf}

\bibitem[{Price et~al(2020)Price, Gifford-Moore, Flemming, Musker, Roichman,
  Sylvain, Thain, Dixon, and Sorensen}]{price-etal-2020-six}
Price I, Gifford-Moore J, Flemming J, et~al (2020) Six attributes of unhealthy
  conversations. In: Proceedings of the Fourth Workshop on Online Abuse and
  Harms. Association for Computational Linguistics, Online, pp 114--124,
  \doi{10.18653/v1/2020.alw-1.15},
  \urlprefix\url{https://aclanthology.org/2020.alw-1.15}

\bibitem[{Rajpurkar et~al(2016)Rajpurkar, Zhang, Lopyrev, and
  Liang}]{rajpurkar-etal-2016-squad}
Rajpurkar P, Zhang J, Lopyrev K, et~al (2016) {SQ}u{AD}: 100,000+ questions for
  machine comprehension of text. In: Proceedings of the 2016 Conference on
  Empirical Methods in Natural Language Processing. Association for
  Computational Linguistics, Austin, Texas, pp 2383--2392,
  \doi{10.18653/v1/D16-1264}, \urlprefix\url{https://aclanthology.org/D16-1264}

\bibitem[{Rosen(2010)}]{inbook}
Rosen A (2010) Morphological Tags in Parallel Corpora, pp 205--234

\bibitem[{Schwenk and Li(2018)}]{schwenk-li-2018-corpus}
Schwenk H, Li X (2018) A corpus for multilingual document classification in
  eight languages. In: Proceedings of the Eleventh International Conference on
  Language Resources and Evaluation ({LREC} 2018). European Language Resources
  Association (ELRA), Miyazaki, Japan,
  \urlprefix\url{https://aclanthology.org/L18-1560}

\bibitem[{Tiedemann(2012)}]{TIEDEMANN12.463}
Tiedemann J (2012) Parallel data, tools and interfaces in {OPUS}. In: Chair)
  NCC, Choukri K, Declerck T, et~al (eds) Proceedings of the Eight
  International Conference on Language Resources and Evaluation (LREC'12).
  European Language Resources Association (ELRA), Istanbul, Turkey

\bibitem[{Williams et~al(2018)Williams, Nangia, and Bowman}]{N18-1101}
Williams A, Nangia N, Bowman S (2018) A broad-coverage challenge corpus for
  sentence understanding through inference. In: Proceedings of the 2018
  Conference of the North American Chapter of the Association for Computational
  Linguistics: Human Language Technologies, Volume 1 (Long Papers). Association
  for Computational Linguistics, pp 1112--1122,
  \urlprefix\url{http://aclweb.org/anthology/N18-1101}

\bibitem[{Ziemski et~al(2016)Ziemski, Junczys-Dowmunt, and
  Pouliquen}]{ziemski-etal-2016-united}
Ziemski M, Junczys-Dowmunt M, Pouliquen B (2016) The {U}nited {N}ations
  parallel corpus v1.0. In: Proceedings of the Tenth International Conference
  on Language Resources and Evaluation ({LREC}'16). European Language Resources
  Association (ELRA), Portoro{\v{z}}, Slovenia, pp 3530--3534,
  \urlprefix\url{https://aclanthology.org/L16-1561}

\end{thebibliography}

\begin{appendices}

\section{Annotation interface}
\label{sec:annotation_interface}

\begin{figure}[h!]
\centering
\includegraphics[width=10cm, height=4cm]{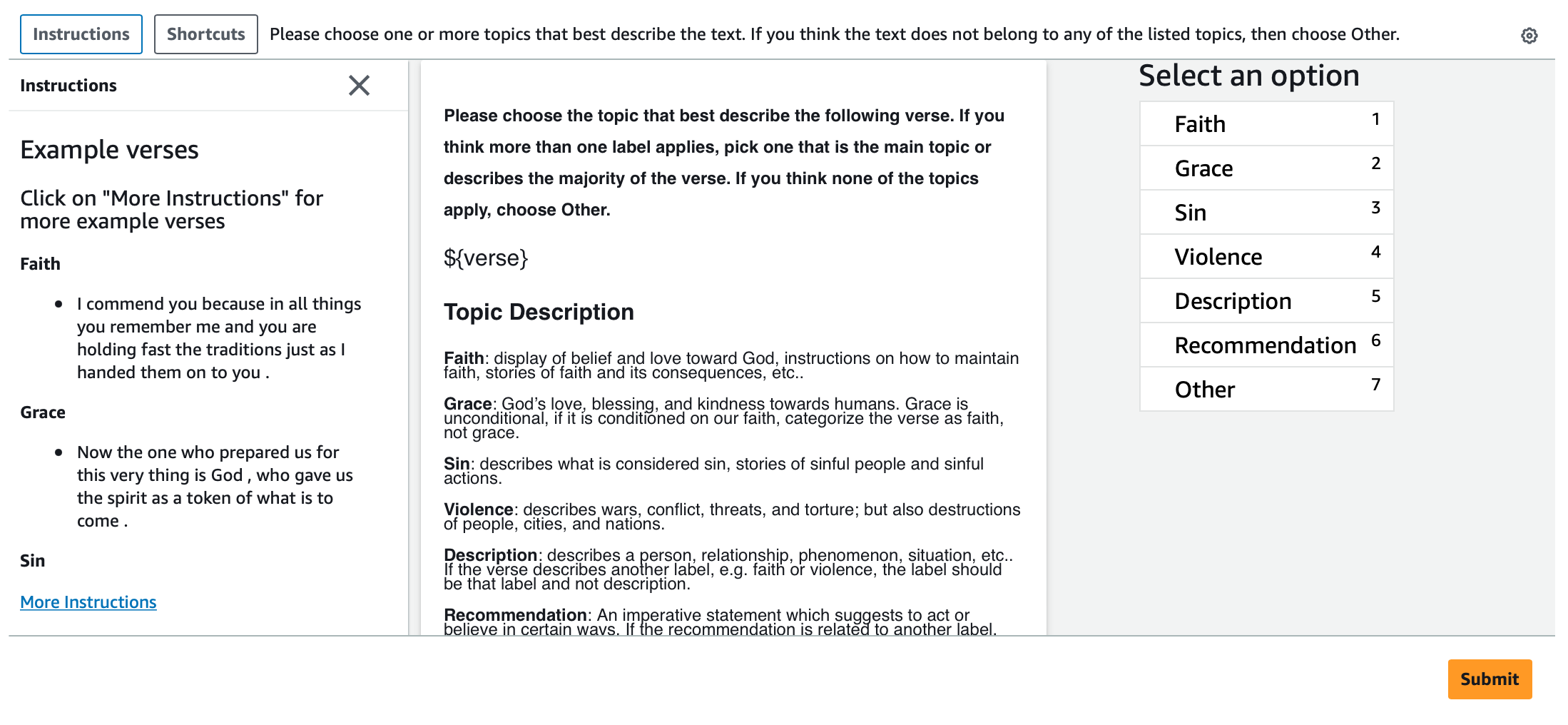}
\caption{mTurk interface with English instructions and verse examples}
\label{fig:mturk_screenshot}
\end{figure}

Figure \ref{fig:mturk_screenshot} shows a screenshot of the annotation interface. Workers are asked to select one label for each verse among six labels. If they think one verse does not belong to any of them, the workers should classify this verse into Other.

\section{Data collection}
Our dataset is built based on PBC and 1000Langs. Due to the copyright issue, our dataset consists of three parts:
\begin{itemize}
    \item 1403 editions in 670 languages with permissive
      licenses which we distribute freely  (the corpus we
      call Taxi1500-c v1.0).
    \item For the remaining PBC Bibles, please contact
      Michael Cysouw at Philipps University of Marburg to
      request access to PBC. Once granted access, run the code available at \url{https://github.com/cisnlp/Taxi1500/corpus_obtain} to obtain the labeled dataset.
    \item For the remaining 1000Langs Bibles, use the code provided at \url{https://github.com/ehsanasgari/1000Langs} to crawl the corpus. Then, run the code available at \url{https://github.com/cisnlp/Taxi1500/corpus_obtain} to obtain the labeled dataset.
\end{itemize}

\section{Results for zero-shot}
\label{sec:appendix_zero_shot_results}
We report the detailed results for zero-shot transfer of BOW, mBERT, XLM-R-B, XLM-R-L, and Glot500-m.

\begin{table*}
\centering
    \resizebox{\textwidth}{!}{
\begin{tabular}{cccccc|cccccc}
    \toprule
lan\_script & BOW & mBert & XLM-R-B & XLM-R-L & Glot500-m & lan\_script & BOW & mBert & XLM-R-B & XLM-R-L & Glot500-m \\
aah\_Latn &\textbf{0.13} &0.10 &0.05 &0.05 &0.08&aoz\_Latn &\textbf{0.21} &0.13 &0.07 &0.05 &0.07\\
aai\_Latn &\textbf{0.22} &0.15 &0.09 &0.05 &0.09&apb\_Latn &0.07 &0.08 &0.06 &0.05 &\textbf{0.12}\\
aak\_Latn &0.07 &\textbf{0.13} &0.05 &0.05 &0.05&ape\_Latn &0.13 &\textbf{0.13} &0.05 &0.05 &0.07\\
aau\_Latn &0.12 &\textbf{0.12} &0.06 &0.05 &0.10&apn\_Latn &0.07 &\textbf{0.19} &0.06 &0.05 &0.05\\
aaz\_Latn &0.07 &\textbf{0.12} &0.05 &0.05 &0.08&apr\_Latn &0.07 &\textbf{0.07} &0.07 &0.05 &0.05\\
abi\_Latn &0.07 &\textbf{0.11} &0.05 &0.05 &0.05&apt\_Latn &0.08 &\textbf{0.14} &0.07 &0.05 &0.07\\
abt\_Latn &0.09 &\textbf{0.13} &0.08 &0.05 &0.06&apu\_Latn &0.07 &0.09 &\textbf{0.10} &0.05 &0.05\\
\bottomrule
\end{tabular}
    }
   \caption{zero-shot score of BOW, mBERT, XLM-R-B, XLM-R-L, and Glot500-m.}
    \label{tab:zero_shot_table1}
\end{table*}

\begin{table*}
\centering
    \resizebox{\textwidth}{!}{
\begin{tabular}{cccccc|cccccc}
    \toprule
lan\_script & BOW & mBert & XLM-R-B & XLM-R-L & Glot500-m & lan\_script & BOW & mBert & XLM-R-B & XLM-R-L & Glot500-m \\
bgz\_Latn &0.09 &\textbf{0.18} &0.09 &0.06 &0.15&bzj\_Latn &0.24 &0.15 &0.13 &0.06 &\textbf{0.35}\\
bhl\_Latn &0.10 &\textbf{0.12} &0.06 &0.05 &0.07&caa\_Latn &0.14 &\textbf{0.15} &0.07 &0.05 &0.12\\
bhp\_Latn &0.09 &0.11 &\textbf{0.16} &0.06 &0.09&cab\_Latn &0.07 &\textbf{0.10} &0.05 &0.05 &0.05\\
bhw\_Latn &0.09 &\textbf{0.16} &0.07 &0.05 &0.14&cac\_Latn &0.12 &0.12 &0.06 &0.05 &\textbf{0.21}\\
bhz\_Latn &\textbf{0.18} &0.14 &0.06 &0.05 &0.06&caf\_Latn &\textbf{0.09} &0.07 &0.05 &0.05 &0.05\\
bib\_Latn &\textbf{0.16} &0.06 &0.05 &0.05 &0.06&cag\_Latn &0.07 &\textbf{0.14} &0.05 &0.05 &0.11\\
big\_Latn &0.09 &\textbf{0.10} &0.05 &0.05 &0.05&cak\_Latn &0.04 &0.12 &0.05 &0.05 &\textbf{0.42}\\
\bottomrule
\end{tabular}
    }
   \caption{zero-shot score of BOW, mBERT, XLM-R-B, XLM-R-L, and Glot500-m.}
    \label{tab:zero_shot_table1}
\end{table*}

\begin{table*}
\centering
    \resizebox{\textwidth}{!}{
\begin{tabular}{cccccc|cccccc}
    \toprule
lan\_script & BOW & mBert & XLM-R-B & XLM-R-L & Glot500-m & lan\_script & BOW & mBert & XLM-R-B & XLM-R-L & Glot500-m \\
cpa\_Latn &0.07 &\textbf{0.11} &0.05 &0.05 &0.05&due\_Latn &0.10 &0.12 &0.16 &0.05 &\textbf{0.20}\\
cpb\_Latn &0.07 &\textbf{0.08} &0.08 &0.05 &0.05&dug\_Latn &0.08 &\textbf{0.17} &0.17 &0.11 &0.16\\
cpc\_Latn &0.09 &\textbf{0.12} &0.06 &0.05 &0.05&duo\_Latn &0.14 &0.08 &0.16 &0.06 &\textbf{0.31}\\
cpu\_Latn &0.09 &\textbf{0.11} &0.04 &0.07 &0.05&dur\_Latn &0.10 &\textbf{0.10} &0.05 &0.05 &0.05\\
cpy\_Latn &0.07 &\textbf{0.08} &0.05 &0.05 &0.05&dwr\_Latn &\textbf{0.15} &0.11 &0.06 &0.05 &0.10\\
crh\_Cyrl &0.07 &0.19 &0.15 &0.20 &\textbf{0.45}&dww\_Latn &0.07 &0.07 &\textbf{0.08} &0.05 &0.06\\
crj\_Latn &\textbf{0.15} &0.10 &0.05 &0.05 &0.05&dyi\_Latn &\textbf{0.16} &0.13 &0.07 &0.05 &0.06\\
\bottomrule
\end{tabular}
    }
   \caption{zero-shot score of BOW, mBERT, XLM-R-B, XLM-R-L, and Glot500-m.}
    \label{tab:zero_shot_table1}
\end{table*}

\begin{table*}
\centering
    \resizebox{\textwidth}{!}{
\begin{tabular}{cccccc|cccccc}
    \toprule
lan\_script & BOW & mBert & XLM-R-B & XLM-R-L & Glot500-m & lan\_script & BOW & mBert & XLM-R-B & XLM-R-L & Glot500-m \\
gfk\_Latn &\textbf{0.17} &0.12 &0.07 &0.05 &0.10&hlt\_Latn &0.09 &\textbf{0.09} &0.05 &0.05 &0.06\\
ghe\_Deva &0.07 &0.11 &0.20 &0.15 &\textbf{0.28}&hmo\_Latn &0.09 &\textbf{0.14} &0.09 &0.05 &0.07\\
ghs\_Latn &0.07 &\textbf{0.10} &0.05 &0.05 &0.06&hmr\_Latn &\textbf{0.21} &0.06 &0.07 &0.05 &0.20\\
gid\_Latn &\textbf{0.10} &0.05 &0.05 &0.05 &0.08&hne\_Deva &0.07 &0.27 &0.29 &0.39 &\textbf{0.60}\\
gil\_Latn &0.07 &0.08 &0.04 &0.05 &\textbf{0.23}&hnj\_Latn &0.06 &\textbf{0.06} &0.06 &0.05 &0.05\\
giz\_Latn &0.07 &\textbf{0.14} &0.06 &0.05 &0.07&hnn\_Latn &0.11 &0.17 &0.17 &0.12 &\textbf{0.31}\\
gjn\_Latn &0.09 &\textbf{0.13} &0.05 &0.05 &0.05&hns\_Latn &0.13 &0.12 &0.14 &0.12 &\textbf{0.19}\\
\bottomrule
\end{tabular}
    }
   \caption{zero-shot score of BOW, mBERT, XLM-R-B, XLM-R-L, and Glot500-m.}
    \label{tab:zero_shot_table1}
\end{table*}

\begin{table*}
\centering
    \resizebox{\textwidth}{!}{
\begin{tabular}{cccccc|cccccc}
    \toprule
lan\_script & BOW & mBert & XLM-R-B & XLM-R-L & Glot500-m & lan\_script & BOW & mBert & XLM-R-B & XLM-R-L & Glot500-m \\
jvn\_Latn &0.07 &0.35 &0.36 &\textbf{0.52} &0.49&knf\_Latn &0.13 &\textbf{0.15} &0.07 &0.05 &0.05\\
kaa\_Cyrl &0.07 &0.17 &0.14 &0.16 &\textbf{0.52}&kng\_Latn &0.07 &0.14 &0.08 &0.05 &\textbf{0.15}\\
kab\_Latn &0.11 &\textbf{0.14} &0.07 &0.06 &0.13&knj\_Latn &0.07 &0.09 &0.05 &0.05 &\textbf{0.18}\\
kac\_Latn &\textbf{0.13} &0.10 &0.05 &0.05 &0.05&knk\_Latn &0.06 &\textbf{0.11} &0.05 &0.05 &0.08\\
kal\_Latn &0.09 &0.11 &0.05 &0.05 &\textbf{0.13}&kno\_Latn &0.10 &\textbf{0.10} &0.05 &0.05 &0.07\\
kan\_Knda &0.07 &0.34 &0.56 &\textbf{0.64} &0.61&knv\_Latn &\textbf{0.18} &0.12 &0.05 &0.05 &0.08\\
kao\_Latn &0.09 &\textbf{0.09} &0.05 &0.05 &0.06&kog\_Latn &0.11 &\textbf{0.12} &0.06 &0.05 &0.05\\
\bottomrule
\end{tabular}
    }
   \caption{zero-shot score of BOW, mBERT, XLM-R-B, XLM-R-L, and Glot500-m.}
    \label{tab:zero_shot_table1}
\end{table*}

\begin{table*}
\centering
    \resizebox{\textwidth}{!}{
\begin{tabular}{cccccc|cccccc}
    \toprule
lan\_script & BOW & mBert & XLM-R-B & XLM-R-L & Glot500-m & lan\_script & BOW & mBert & XLM-R-B & XLM-R-L & Glot500-m \\
lap\_Latn &0.14 &\textbf{0.15} &0.06 &0.05 &0.08&mbb\_Latn &0.11 &\textbf{0.20} &0.10 &0.05 &0.10\\
las\_Latn &0.09 &\textbf{0.09} &0.05 &0.05 &0.05&mbc\_Latn &0.12 &\textbf{0.13} &0.05 &0.05 &0.05\\
lat\_Latn &0.14 &0.30 &0.55 &\textbf{0.62} &0.56&mbd\_Latn &\textbf{0.13} &0.12 &0.11 &0.05 &0.10\\
lav\_Latn &0.08 &0.34 &\textbf{0.62} &0.55 &0.52&mbf\_Latn &0.07 &0.31 &0.49 &\textbf{0.57} &0.56\\
law\_Latn &0.09 &\textbf{0.09} &0.06 &0.05 &0.09&mbh\_Latn &0.15 &\textbf{0.15} &0.07 &0.05 &0.09\\
lbk\_Latn &0.12 &0.10 &0.09 &0.05 &\textbf{0.14}&mbi\_Latn &0.13 &\textbf{0.17} &0.08 &0.05 &0.06\\
lcm\_Latn &0.16 &\textbf{0.20} &0.05 &0.06 &0.15&mbj\_Latn &\textbf{0.16} &0.14 &0.08 &0.05 &0.06\\
\bottomrule
\end{tabular}
    }
   \caption{zero-shot score of BOW, mBERT, XLM-R-B, XLM-R-L, and Glot500-m.}
    \label{tab:zero_shot_table1}
\end{table*}

\begin{table*}
\centering
    \resizebox{\textwidth}{!}{
\begin{tabular}{cccccc|cccccc}
    \toprule
lan\_script & BOW & mBert & XLM-R-B & XLM-R-L & Glot500-m & lan\_script & BOW & mBert & XLM-R-B & XLM-R-L & Glot500-m \\
mlt\_Latn &0.11 &0.16 &0.05 &0.06 &\textbf{0.29}&mzm\_Latn &0.09 &\textbf{0.09} &0.05 &0.05 &0.05\\
mmn\_Latn &0.17 &0.19 &0.18 &0.21 &\textbf{0.32}&mzw\_Latn &0.05 &\textbf{0.09} &0.05 &0.05 &0.06\\
mmo\_Latn &\textbf{0.17} &0.09 &0.09 &0.05 &0.05&nab\_Latn &0.07 &\textbf{0.14} &0.05 &0.05 &0.05\\
mmx\_Latn &\textbf{0.14} &0.11 &0.05 &0.05 &0.06&naf\_Latn &0.07 &\textbf{0.15} &0.05 &0.05 &0.06\\
mna\_Latn &\textbf{0.11} &0.08 &0.05 &0.05 &0.05&nak\_Latn &0.11 &\textbf{0.12} &0.04 &0.05 &0.08\\
mnb\_Latn &0.10 &\textbf{0.17} &0.06 &0.05 &0.16&nan\_Latn &\textbf{0.14} &0.11 &0.05 &0.05 &0.06\\
mnf\_Latn &0.11 &\textbf{0.13} &0.05 &0.05 &0.06&naq\_Latn &0.09 &\textbf{0.10} &0.05 &0.05 &0.07\\
\bottomrule
\end{tabular}
    }
   \caption{zero-shot score of BOW, mBERT, XLM-R-B, XLM-R-L, and Glot500-m.}
    \label{tab:zero_shot_table1}
\end{table*}

\begin{table*}
\centering
    \resizebox{\textwidth}{!}{
\begin{tabular}{cccccc|cccccc}
    \toprule
lan\_script & BOW & mBert & XLM-R-B & XLM-R-L & Glot500-m & lan\_script & BOW & mBert & XLM-R-B & XLM-R-L & Glot500-m \\
nng\_Latn &0.07 &\textbf{0.09} &0.07 &0.05 &0.06&oym\_Latn &0.07 &\textbf{0.12} &0.05 &0.05 &0.05\\
nnh\_Latn &0.08 &\textbf{0.14} &0.07 &0.05 &0.08&ozm\_Latn &\textbf{0.13} &0.06 &0.06 &0.05 &0.05\\
nnl\_Latn &0.12 &\textbf{0.12} &0.07 &0.05 &0.06&pab\_Latn &\textbf{0.12} &0.05 &0.05 &0.05 &0.05\\
nno\_Latn &0.15 &0.46 &\textbf{0.58} &0.56 &0.43&pad\_Latn &0.13 &\textbf{0.15} &0.06 &0.05 &0.06\\
nnp\_Latn &0.07 &\textbf{0.08} &0.07 &0.05 &0.05&pag\_Latn &0.14 &0.14 &0.20 &0.17 &\textbf{0.33}\\
nnq\_Latn &0.14 &\textbf{0.15} &0.11 &0.10 &0.14&pah\_Latn &0.09 &\textbf{0.15} &0.06 &0.05 &0.05\\
nnw\_Latn &\textbf{0.07} &0.05 &0.05 &0.05 &0.05&pam\_Latn &0.13 &0.18 &0.11 &0.11 &\textbf{0.38}\\
\bottomrule
\end{tabular}
    }
   \caption{zero-shot score of BOW, mBERT, XLM-R-B, XLM-R-L, and Glot500-m.}
    \label{tab:zero_shot_table1}
\end{table*}

\begin{table*}
\centering
    \resizebox{\textwidth}{!}{
\begin{tabular}{cccccc|cccccc}
    \toprule
lan\_script & BOW & mBert & XLM-R-B & XLM-R-L & Glot500-m & lan\_script & BOW & mBert & XLM-R-B & XLM-R-L & Glot500-m \\
quw\_Latn &0.07 &0.10 &0.07 &0.05 &\textbf{0.18}&shp\_Latn &0.07 &\textbf{0.12} &0.06 &0.05 &0.05\\
quy\_Latn &0.07 &0.11 &0.07 &0.06 &\textbf{0.27}&shu\_Latn &0.09 &\textbf{0.20} &0.16 &0.11 &0.19\\
quz\_Latn &0.07 &0.10 &0.07 &0.05 &\textbf{0.24}&sig\_Latn &\textbf{0.13} &0.08 &0.05 &0.05 &0.05\\
qva\_Latn &0.07 &0.10 &0.07 &0.05 &\textbf{0.18}&sil\_Latn &\textbf{0.14} &0.07 &0.05 &0.05 &0.05\\
qvc\_Latn &0.09 &\textbf{0.11} &0.06 &0.05 &0.05&sim\_Latn &0.08 &\textbf{0.10} &0.06 &0.05 &0.07\\
qve\_Latn &0.09 &0.13 &0.06 &0.05 &\textbf{0.33}&sin\_Sinh &0.07 &0.16 &0.51 &\textbf{0.67} &0.57\\
qvh\_Latn &0.12 &0.12 &0.05 &0.07 &\textbf{0.24}&sja\_Latn &0.10 &\textbf{0.10} &0.05 &0.05 &0.05\\
\bottomrule
\end{tabular}
    }
   \caption{zero-shot score of BOW, mBERT, XLM-R-B, XLM-R-L, and Glot500-m.}
    \label{tab:zero_shot_table1}
\end{table*}

\begin{table*}
\centering
    \resizebox{\textwidth}{!}{
\begin{tabular}{cccccc|cccccc}
    \toprule
lan\_script & BOW & mBert & XLM-R-B & XLM-R-L & Glot500-m & lan\_script & BOW & mBert & XLM-R-B & XLM-R-L & Glot500-m \\
taq\_Latn &0.10 &\textbf{0.11} &0.07 &0.05 &0.06&tro\_Latn &\textbf{0.15} &0.12 &0.07 &0.05 &0.07\\
tar\_Latn &0.10 &\textbf{0.10} &0.05 &0.05 &0.05&trp\_Latn &\textbf{0.10} &0.08 &0.06 &0.05 &0.05\\
tat\_Cyrl &0.07 &0.31 &0.12 &0.15 &\textbf{0.45}&trq\_Latn &0.05 &\textbf{0.12} &0.05 &0.05 &0.07\\
tav\_Latn &\textbf{0.13} &0.11 &0.05 &0.05 &0.09&trs\_Latn &0.06 &\textbf{0.10} &0.07 &0.05 &0.10\\
taw\_Latn &\textbf{0.14} &0.09 &0.07 &0.05 &0.07&tsg\_Latn &0.11 &0.17 &0.15 &0.11 &\textbf{0.27}\\
tbc\_Latn &0.09 &\textbf{0.12} &0.05 &0.05 &0.06&tsn\_Latn &0.12 &0.12 &0.09 &0.05 &\textbf{0.23}\\
tbg\_Latn &0.07 &\textbf{0.14} &0.08 &0.05 &0.06&tsw\_Latn &0.07 &\textbf{0.12} &0.07 &0.05 &0.08\\
\bottomrule
\end{tabular}
    }
   \caption{zero-shot score of BOW, mBERT, XLM-R-B, XLM-R-L, and Glot500-m.}
    \label{tab:zero_shot_table1}
\end{table*}

\begin{table*}
\centering
    \resizebox{\textwidth}{!}{
\begin{tabular}{cccccc|cccccc}
    \toprule
lan\_script & BOW & mBert & XLM-R-B & XLM-R-L & Glot500-m & lan\_script & BOW & mBert & XLM-R-B & XLM-R-L & Glot500-m \\
wer\_Latn &0.09 &\textbf{0.15} &0.05 &0.05 &0.05&zac\_Latn &0.12 &\textbf{0.20} &0.09 &0.09 &0.18\\
whk\_Latn &0.11 &\textbf{0.17} &0.07 &0.05 &0.11&zad\_Latn &\textbf{0.15} &0.10 &0.04 &0.05 &0.05\\
wim\_Latn &0.07 &\textbf{0.08} &0.06 &0.05 &0.08&zae\_Latn &\textbf{0.14} &0.13 &0.10 &0.05 &0.06\\
wiu\_Latn &0.12 &\textbf{0.13} &0.05 &0.06 &0.05&zai\_Latn &0.08 &0.21 &0.13 &0.09 &\textbf{0.25}\\
wmw\_Latn &0.14 &0.16 &0.23 &0.31 &\textbf{0.41}&zam\_Latn &0.09 &\textbf{0.16} &0.07 &0.05 &0.13\\
wnc\_Latn &0.07 &\textbf{0.12} &0.07 &0.06 &0.05&zao\_Latn &\textbf{0.14} &0.09 &0.06 &0.05 &0.06\\
wnu\_Latn &0.11 &\textbf{0.13} &0.05 &0.05 &0.05&zar\_Latn &0.11 &\textbf{0.17} &0.06 &0.05 &0.08\\
\bottomrule
\end{tabular}
    }
   \caption{zero-shot score of BOW, mBERT, XLM-R-B, XLM-R-L, and Glot500-m.}
    \label{tab:zero_shot_table1}
\end{table*}

\end{appendices}



\end{document}